\documentclass[letterpaper]{article} % DO NOT CHANGE THIS
\usepackage{aaai2027}  % preprint mode: prints the author block
\usepackage[hyphens]{url}  % DO NOT CHANGE THIS
\usepackage{graphicx} % DO NOT CHANGE THIS
\usepackage{natbib}  % DO NOT CHANGE THIS
\usepackage{caption} % DO NOT CHANGE THIS
\usepackage{algorithm}
\usepackage{algorithmic}

\usepackage{amsmath}
\usepackage{amssymb}
\usepackage{mathtools}
\usepackage{amsthm}
\usepackage{booktabs}
\usepackage{hhline}
\usepackage{array}
\usepackage{tabularx}
\usepackage{multirow}
\usepackage{threeparttable}
\usepackage{xcolor}
\usepackage{pifont}
\usepackage{xspace}
\newcommand{\cnum}[1]{\ding{\numexpr171+#1\relax}}
\usepackage[capitalize,noabbrev]{cleveref}
\crefname{appendix}{appendix}{appendices}
\Crefname{appendix}{Appendix}{Appendices}

\newcommand{\rope}{\mathrm{RoPE}}

\newcommand{\A}{\mathbf{A}}
\newcommand{\B}{\mathbf{B}}
\newcommand{\E}{\mathbf{E}}
\newcommand{\W}{\mathbf{W}}
\newcommand{\X}{\mathbf{X}}

\newcommand{\K}{\mathbf{K}}
\newcommand{\Q}{\mathbf{Q}}
\newcommand{\V}{\mathbf{V}}
\newcommand{\KV}{\ensuremath{\K\V}}

\newcommand{\M}{\mathbf{M}}
\newcommand{\R}{\mathbf{R}}

\newcommand{\ropepair}{\mathcal{P}}

\newcommand{\gdn}{{\color{green!55!black}$\downarrow$}}
\newcommand{\rup}{{\color{red}$\uparrow$}}
\definecolor{rapsand}{HTML}{F2C57C}
\newcommand{\sandswatch}{\textcolor{rapsand}{\rule{1.0em}{0.55em}}}
\newcolumntype{Y}{>{\raggedright\arraybackslash}X}

\theoremstyle{plain}

\newtheorem{proposition}{Proposition}

\newtheorem{corollary}{Corollary}
\theoremstyle{definition}

\title{RAP: RoPE-Aligned Pruning for \KV{}-Cache Compression}
\author{
    Jihao Xin,
    Tian Lyu,
    David Keyes,
    Hatem Ltaief,
    Marco Canini
}
\affiliations{
    King Abdullah University of Science and Technology (KAUST)\\
}

\nocopyright   % preprint: the paper is under review, so no AAAI copyright is asserted

\begin{document}

\maketitle

\begin{abstract}
Long-context inference in large language models (LLMs) is bottlenecked by the memory and compute of the key--value (\KV{}) cache.
Structured pruning is a direct way to shrink it: dropping the least useful channels of the $\W_k,\W_v$ projection weights to reduce the output \KV{} dimensions.
However, modern LLMs apply Rotary Position Embedding (RoPE) after the $\Q/\K$ projections, which rotates feature dimensions in pairs. Therefore, removing individual channels breaks these pairs, corrupting RoPE's positional semantics and rendering the pruned model unusable.
We propose \emph{RoPE-Aligned Pruning (RAP)}, which constrains the pruning granularity to \emph{RoPE-aligned pairs} rather than individual channels: removing whole pairs to keep the rotation intact.
Our evaluation across Llama, Mistral, and Qwen models from 3B to 14B shows that RAP maintains usable accuracy at a $30\%$ \KV{} compression ratio, far outperforms RoPE-blind channel pruning, stays near the strongest low-rank method at lower attention cost, and composes with 4-bit weight quantization.
\end{abstract}

\section{Introduction}
\label{sec:intro}
Large Language Models (LLMs)~\citep{achiam2023gpt, grattafiori2024llama, bai2023qwen, liu2024deepseek} have advanced rapidly through scaling model size~\citep{gholami2024ai} and context length.
Yet long-context inference remains bottlenecked by the ever-expanding \KV{} cache: a 30B-scale model at 100K context consumes 22.8~GB solely for the \KV{} cache~\citep{kvcachechallenge}, motivating effective compression for deployment on commodity hardware.

A direct way to shrink the cache is \emph{structured pruning}: dropping the least useful channels of the $\W_k,\W_v$ projection weights to reduce the output \KV{} dimensions, as well as the parameters and FLOPs.
Existing methods fall into two classes by granularity.
\textbf{Per-head} pruning~\citep{llm-pruner} drops whole attention heads, but its coarse granularity quickly destroys quality at high compression.
\textbf{Per-channel} pruning~\citep{kv-pruner} instead prunes $\W_k,\W_v$ channel by channel, retaining important ones at finer budgets.
However, existing channel-pruning methods have often overlooked the pair structure imposed by Rotary Position Embedding (\textbf{RoPE})~\citep{rope} (\Cref{app:rope_adoption}).
Because tensor shapes remain valid, this failure can be silent: execution raises no error, and the mismatch appears only as lower accuracy.
Existing pruning methods impose a post-hoc retraining step to adapt around the mismatch and regain substantial accuracy, further masking the cause without restoring the removed partner.
Low-rank methods decompose $\W_k$ into $\A_k$ and $\B_k$ to cache a low-dimensional latent $\X\A_k$~\citep{svd-llm, palu}, but RoPE prevents its expansion $\B_k$ from fusing into $\W_q$ and forces full-$\K$ reconstruction at every decode step (\Cref{sec:motivation}).

\begin{figure*}[!t]
    \centering
    \includegraphics[width=0.86\textwidth]{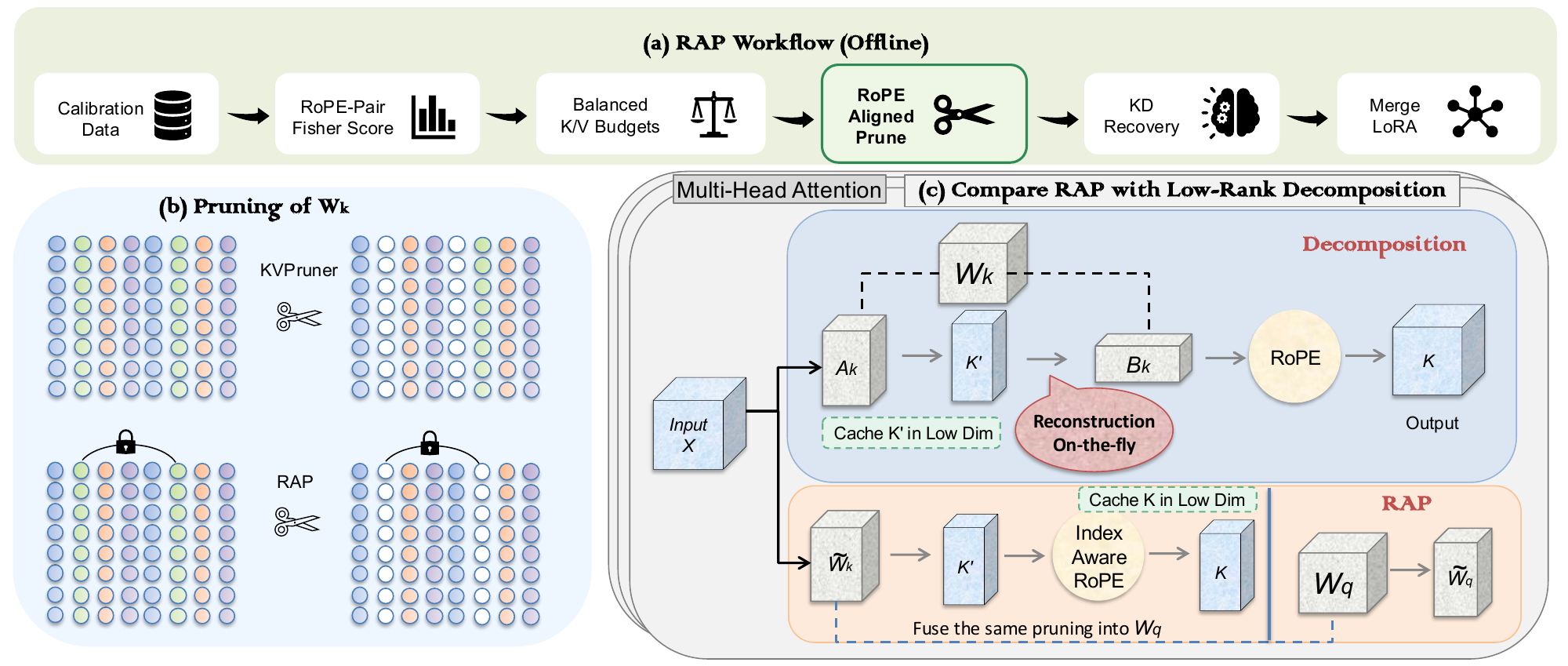}
    \caption{RAP overview. (a) Offline pipeline of RAP. (b) Channel pruning orphans half-pairs, while RAP removes whole pairs. (c) Low-rank reconstructs $\K$ each step, while RAP runs on pruned dimensions after fusing the pruning into $\widetilde{\W}_q$.}
    \label{fig:rap_overview}
\end{figure*}

We propose \emph{RoPE-Aligned Pruning (RAP)}, which prunes at RoPE-pair granularity wherever RoPE is applied (\Cref{fig:rap_overview}).
In most modern LLMs, RoPE acts on the $\Q/\K$ path, so RAP ranks $\W_k$'s RoPE pairs by Fisher scores on a small calibration set, keeps the top pairs under a target retain ratio, and prunes $\W_v$ by channels. The pair-aligned cut then fuses into $\W_q$ offline, so $\W_q$ shrinks to the same low dimension and decode needs no per-step reconstruction (\Cref{sec:rap}).
As a result, RAP can cut about 10\% of the \KV{} dimensions ($\rho{=}0.9$) without fine-tuning, while $30\%$ compression ($\rho{=}0.7$) recovers accuracy with lightweight LoRA knowledge distillation (KD). Existing pruning often needs heavy retraining, and decomposition-based methods pay a reconstruction tax at every decode step.
Unlike activation-side compressors that only shrink the cached tensors, RAP is \emph{weight-side} pruning, so the \KV{} cache, parameters, and FLOPs shrink together.
RAP is also orthogonal to \KV{} quantization which we evaluate in \Cref{sec:experiments}.

We evaluate on Llama, Mistral, and Qwen models from 3B to 14B, reporting WikiText-2 perplexity, zero-shot accuracy, and LongBench~\citep{longbench}.
Baselines include KVPruner~\citep{kv-pruner}, PaLU~\citep{palu}, and SVD-LLM~\citep{svd-llm}. The appendix further compares index-aware KVPruner, whole-head pruning, EliteKV, and MatryoshkaKV (\Cref{app:baselines}).
At iso-\KV{} (matched retain ratio $\rho$), RAP is the only accurate method whose \KV{} cache, parameters, and FLOPs co-scale with $\rho$ without reconstruction (\Cref{tab:rap_complexity}).
It keeps every RoPE pair intact by construction, so it stays far ahead of KVPruner under the same KD, and stays close to PaLU while leading on LongBench once attention compute is matched (\Cref{tab:main_results,app:efficiency}).

\section{Background}
\label{sec:background}
\subsection{Transformers and the \KV{} Cache}
Modern LLMs are decoder-only transformers built on multi-head attention (MHA)~\citep{attention}.
Let $H$ be the number of attention heads and $\hat{D}$ the model dimension, so each head has dimension $D=\hat{D}/H$.
At decoding step $t$, a layer receives $X_t$ and forms query, key, and value states
\begin{equation}
    \Q_t = X_t\W_q,\qquad
    \K_t = X_t\W_k,\qquad
    \V_t = X_t\W_v,
    \label{eq:qkv-proj}
\end{equation}
with per-head projections $\W_q,\W_k,\W_v\in\mathbb{R}^{\hat{D}\times D}$.
Attention over the prefix $1{:}t$ is then
\begin{equation}
    \mathrm{Attn}(\Q_t,\K_{1:t},\V_{1:t})
    =
    \mathrm{softmax}\!\left(\frac{\Q_t\K_{1:t}^\top}{\sqrt{D}}\right)\V_{1:t},
    \label{eq:attn}
\end{equation}
and the concatenation of all heads is mapped back by an output projection $\W_o\in\mathbb{R}^{\hat{D}\times\hat{D}}$.
Grouped-Query Attention (GQA) uses the same form with fewer \KV{} heads shared across $\Q$ heads. Our notation below uses one head for clarity.

Autoregressive decoding writes each new $\K_t,\V_t$ into a \textbf{cache} and reuses it later, so past projections need not be recomputed.
Cache size grows as $O(S)$ with length $S$, while attention compute scales as $O(S^2)$.
At long contexts, the cache dominates GPU memory and compute, which is the bottleneck that \KV{}-cache compression targets.

\subsection{\KV{}-Cache Compression}
\label{sec:background-kvcache}
To alleviate this bottleneck, prior work compresses the \KV{} cache along different axes.
We organize methods first by \emph{what} they change (object), then by \emph{how} (methodology).

\textbf{By object.}
\emph{Activation-side} methods (ThinK~\citep{think}, SparK~\citep{spark2025}, LeanK~\citep{leank2025}, RotateK~\citep{rotatek2026} (for VLMs), MatryoshkaKV~\citep{matryoshkakv}) mask or project the already-computed $\KV$ at inference.
They shrink the cache directly but leave the pretrained $\W_k,\W_v$ untouched, so parameters and projection FLOPs are not reduced.
\emph{Weight-side} methods change the projection weights, so the parameter size, produced cache, and FLOPs will shrink together.
RAP is weight-side.

\textbf{By methodology.}
\textbf{Quantization}~\citep{kivi, kvquant} lowers the bit-width of cached $\K$ and $\V$.
\textbf{Token eviction}~\citep{snapkv, h2o, kvzip} drops less important tokens from the cache.
\textbf{Low-rank decomposition} (PaLU~\citep{palu}, SVD-LLM~\citep{svd-llm}, EliteKV~\citep{elitekv}, ReCalKV~\citep{recalkv2025}, Thin-Keys~\citep{thin-keys}) writes $\W_k\approx\A\B$ and caches $\X\A$.
These families are orthogonal to RAP and compose with it.
RAP belongs to \textbf{structured pruning}, which removes \KV{} feature dimensions.
Unstructured weight sparsity~\citep{sparsegpt, wanda} is RoPE-safe but leaves the cached \KV{} tensors dense. Head pruning~\citep{llm-pruner} is also RoPE-safe but coarse: the budget moves in whole-head steps, so high compression quickly destroys quality (\Cref{app:baselines}).
Channel pruning~\citep{kv-pruner} is finer, but orphans RoPE half-pairs and breaks positional semantics (\Cref{sec:motivation}).
\Cref{tab:capability} summarizes the trade-offs: RAP is the only method that saves cache, parameters, and FLOPs together while staying reconstruction-free at a high compression ratio.
\begin{table}[t]
    \centering
    \small
    \setlength{\tabcolsep}{3pt}
    \begin{tabular}{l ccccc}
    \toprule
    \textbf{Family} & \shortstack{\textbf{Cache}\\\textbf{Save}} & \shortstack{\textbf{Para.}\\\textbf{Save}} & \shortstack{\textbf{FLOPs}\\\textbf{Save}} & \shortstack{\textbf{Recon.}\\\textbf{Free}} & \shortstack{\textbf{Comp.}\\\textbf{Ratio}} \\
    \midrule
    Activation-side & \checkmark & --- & --- & \checkmark & high \\
    \midrule
    \multicolumn{6}{l}{\emph{Weight-side}} \\
    Low-rank & \checkmark & --- & --- & --- & high \\
    Head pruning & \checkmark & \checkmark & \checkmark & \checkmark & low \\
    Channel pruning & \checkmark & \checkmark & \checkmark & \checkmark & low \\
    \textbf{RAP} & \checkmark & \checkmark & \checkmark & \checkmark & high \\
    \bottomrule
    \end{tabular}
    \caption{Capability comparison of \KV{} compression methods. RAP is the only row that meets all criteria.}
    \label{tab:capability}
\end{table}

We compare methods at \textbf{iso-\KV{}-cache}: the same retain ratio $\rho$ (fraction of dense \KV{} channels kept), so quality and efficiency gaps are not confounded by unequal cache budgets.
The direct baseline is KVPruner~\citep{kv-pruner}, the channel-pruning counterpart of RAP.
We also compare with low-rank compression (PaLU~\citep{palu}, SVD-LLM~\citep{svd-llm}).

\section{Motivation: RoPE Incompatibility}
\label{sec:motivation}
RoPE encodes position by rotating paired feature dimensions of each $\K$ and $\Q$.
At decoding step $t$ (\Cref{sec:background}), a $\K$ pair is rotated as
\begin{equation}
    \rope\left(\begin{bmatrix} k_{t,j} \\ k_{t,j'} \end{bmatrix}\right) = \mathbf{R}(t,\theta_j) \begin{bmatrix} k_{t,j} \\ k_{t,j'} \end{bmatrix},
    \label{eq:rope}
\end{equation}
with $\mathbf{R}(t,\theta_j) = \left[\begin{smallmatrix} \cos(t\theta_j) & -\sin(t\theta_j) \\ \sin(t\theta_j) & \cos(t\theta_j) \end{smallmatrix}\right]$ and $\theta_j = \theta_{\text{base}}^{-2j/D}$.
Here $\theta_{\text{base}}$ is a fixed model constant (often $10^{4}$--$10^{6}$) that controls how fast each pair rotates.
In each head, $\K_t$ has $D$ dims that RoPE splits into $D/2$ \emph{pairs} $(j,j')$.
Each pair is rotated by a $2{\times}2$ block.
Models group dims in two ways: next to each other, $(j,j')=(2x{-}1,2x)$, or half apart, $(j,j')=(x,x{+}D/2)$.
In both cases the two dims of a pair must be kept or removed together.
If a channel pruner keeps one and drops the other, RoPE has no partner left, position information breaks, and the pruned model fails (\Cref{sec:accuracy}).
Crucially, existing work often overlooks this~\citep{kv-pruner}, as breaking a RoPE pair is \textbf{silent}, with no runtime error, yet it is a primary cause of accuracy collapse.

The same pairing also forces per-step reconstruction in weight-side low-rank compression~\citep{palu, thin-keys}.
Low-rank decomposition writes $\W_k \approx \A_k\B_k$ and aims to cache only the low-dimensional latent $\X\A_k$.
Without RoPE, $\B_k$ can be fused with $\W_q$ as $\widetilde\W_q=\W_q \B_k^\top$, so the low-dimensional cache needs no reconstruction.
However, with modern LLMs applying RoPE after the $\Q/\K$ projections, the decomposition does \textbf{not commute with RoPE}: $\rope(\X\A_k\B_k) \neq \rope(\X\A_k)\,\B_k$, so $\B_k$ cannot be fused and one must reconstruct $\K$ to full dimension at runtime, resulting in memory and compute overhead (\Cref{fig:rap_overview}).

To avoid reconstruction, we need an expansion that commutes with RoPE and can fuse into $\W_q$ offline.
RAP keeps the two-factor form of low-rank methods, but builds the expansion by pruning: it drops whole RoPE pairs and writes the kept keys as $\W_k\approx\widetilde\W_k\E_k$, where $\E_k$ is a binary, pair-preserving selection matrix.
Unlike a dense $\B_k$, this $\E_k$ \emph{does} commute with RoPE,
\begin{equation}
    \rope(\X\widetilde\W_k\E_k)=\rope(\X\widetilde\W_k)\,\E_k,
    \label{eq:rope-commute-motivation}
\end{equation}
so $\E_k$ can fuse into $\W_q$ as $\widetilde\W_q=\W_q\E_k^\top$ and decode stays reconstruction-free (\Cref{sec:rap}).
In practice, fusing $\E_k$ means pruning $\W_q$ with the same retained indices as $\W_k$.

\section{RAP: RoPE-Aligned Pruning}
\label{sec:rap}
\Cref{sec:motivation} shows that a RoPE-commuting expansion is what makes reconstruction-free decode possible.
RAP turns that requirement into a pruning pipeline at RoPE-pair granularity.
Taking $\K$ as an example, RAP builds a pruned projection $\widetilde\W_k$ and a binary, pair-preserving expansion $\E_k$ so that RoPE-commutativity holds (\Cref{eq:rope-commute-motivation}).
On the compressed matrices, $\rope(\cdot)$ is an \emph{index-aware} rotation that uses the original dimension indices of the retained pairs.
Thus $\E_k$ fuses into $\W_q$ as $\widetilde\W_q=\W_q\E_k^\top$.
\Cref{alg:rap} summarizes the steps below.

\begin{algorithm}[t]
    \caption{RoPE-Aligned Pruning (RAP)}
    \label{alg:rap}
    \begin{algorithmic}[1]
        \REQUIRE model $\mathcal{M}$ with $L$ layers and $H_{kv}$ \KV{} heads, calibration data $\mathcal{D}$, retain ratio $\rho$
        \ENSURE pruned model $\widetilde{\mathcal{M}}$
        \STATE Collect $\W_q^{(\ell,h)},\W_k^{(\ell,h)},\W_v^{(\ell,h)}$, and $\W_o^{(\ell)}$ from $\mathcal{M}$
        \FOR{$\ell=1$ to $L$}
            \FOR{$h=1$ to $H_{kv}$}
                \STATE Partition $\W_k^{(\ell,h)}$ into RoPE pairs $\ropepair$
                \STATE Compute $\W_k^{(\ell,h)}$ pair Fisher scores $\{\sigma_p^{(\ell,h)}\}_{p\in\ropepair}$
                \STATE Compute $\W_v^{(\ell,h)}$ channel Fisher scores $\{\sigma_i^{(\ell,h)}\}_{i=1}^{D}$
            \ENDFOR
        \ENDFOR
        \STATE \textit{\# \Cref{alg:allocation} returns retained $\K$ pairs and $\V$ channels}
        \STATE $\{\mathcal{S}_k^{(\ell,h)},\mathcal{S}_v^{(\ell,h)}\}\gets\textsc{Budget-Alloc}(\{\sigma_p^{(\ell,h)}\},\{\sigma_i^{(\ell,h)}\},\rho)$
        \FOR{$\ell=1$ to $L$}
            \FOR{$h=1$ to $H_{kv}$}
                \STATE Keep pairs $\mathcal{S}_k^{(\ell,h)}$ of $\W_k^{(\ell,h)}$ as $\widetilde\W_k^{(\ell,h)}$
                \STATE Build binary expansion $\E_k^{(\ell,h)}$ (Eq.~\eqref{eq:pruned-decomposition})
                \STATE $\widetilde\W_q^{(\ell,h)}\gets\W_q^{(\ell,h)}(\E_k^{(\ell,h)})^\top$ \textit{\# fuse into query}
                \STATE Keep channels $\mathcal{S}_v^{(\ell,h)}$ of $\W_v^{(\ell,h)}$ as $\widetilde\W_v^{(\ell,h)}$
                \STATE Build binary expansion $\E_v^{(\ell,h)}$
                \STATE $\widetilde\W_o^{(\ell,h)}\gets\E_v^{(\ell,h)}\W_o^{(\ell,h)}$ \textit{\# fuse into output}
            \ENDFOR
        \ENDFOR
        \STATE Recover with LoRA--KD
        \STATE Merge LoRA adapters to construct $\widetilde{\mathcal{M}}$
        \STATE \textbf{return} $\widetilde{\mathcal{M}}$
    \end{algorithmic}
\end{algorithm}

\subsection{Step 1: RoPE Pair Scoring}
Given a total pruning budget set by the retain ratio $\rho$, the first problem is to decide which RoPE pairs are more important to keep.
For each $\W_k\in\mathcal{W}$, we group its columns into RoPE pairs $\ropepair$ ($|\ropepair|=D/2$) following the model's pairing strategy, so every prune-or-keep decision acts on a whole rotation block.
To measure this importance, we use Fisher information as a standard curvature proxy for saliency-based pruning~\citep{lecun-obd, molchanov2019importance, kwon2022fast}.
On calibration data $\mathcal{D}$ we estimate a diagonal empirical Fisher entrywise,
\begin{equation}
    F_{n,i}
    =
    \mathbb{E}_{(x,y)\sim\mathcal{D}}\!\left[
      \left(\frac{\partial \mathcal{L}(x,y)}{\partial (\W_k)_{n,i}}\right)^2
    \right],
    \label{eq:fisher}
\end{equation}
where $n$ indexes the input features of $\W_k$ and $i$ its output channels.
For a RoPE pair $p=(j,j')$, we score the two channels by the aggregated root-Fisher magnitude
\begin{equation}
    \sigma_p
    =
    \sum_{i\in\{j,j'\}}
    \sum_{n}
    \sqrt{F_{n,i}}\,.
    \label{eq:score}
\end{equation}
This $\sigma_p$ is a cheaper surrogate for the second-order saliency $s_p=\sum_{i\in p}\sum_n F_{n,i}(\W_k)_{n,i}^2$ in \Cref{sec:rap-theory}.
It drops the $(\W_k)_{n,i}^2$ weighting, so its ranking can differ, and we validate the ranking empirically (\Cref{sec:ablation}).
For $\W_v$, which has no RoPE pairing, we score each output channel by the same root-Fisher magnitude, $\sigma_i=\sum_n\sqrt{F_{n,i}}$.

\subsection{Step 2: Budget Allocation}
\label{sec:allocation}
Each layer stores both a $\K$ cache and a $\V$ cache, and each cache is further split across attention heads under multi-head attention (MHA) or grouped-query attention (GQA).
The global retain ratio $\rho$ must therefore become a concrete budget for every layer and every \KV{} head.
RAP sets these budgets with three principles (\Cref{alg:allocation}).
\cnum{1}~\textbf{Layer-adaptive.} Layers are not equally sensitive to pruning~\citep{lethe}. A single-layer probe indicates that early and late layers are more fragile than the middle (\Cref{app:method}), so RAP aggregates pair scores within each layer into $\sigma_\ell$ and gives more sensitive layers a larger budget while keeping the mean retain ratio at $\rho$.
\cnum{2}~\textbf{Head-uniform.} Heads within a layer also differ in importance, yet RAP assigns every head within a layer the same budget $m_\ell$ so inference can use \textit{batched GEMM}. Heads may still keep different RoPE pairs; only the pair count is fixed.
\cnum{3}~\textbf{Balanced $\K$/$\V$.} $\K$ and $\V$ play different roles, so their Fisher scores are not interchangeable budgets. In practice $\V$ carries a larger Fisher mass than $\K$. If we pool $\K$ and $\V$ into one Fisher allocation at $\rho{=}0.5$, $\K$ is crushed to ${\sim}21\%$ of the head dimension and long context suffers (\Cref{sec:ablation}). RAP therefore gives $\K$ and $\V$ the same per-head channel budget: $\K$ keeps $m_\ell$ pairs and $\V$ keeps $2m_\ell$ channels.

\begin{algorithm}[t]
    \caption{Budget-Alloc}
    \label{alg:allocation}
    \begin{algorithmic}[1]
        \REQUIRE pair scores $\{\sigma_p^{(\ell,h)}\}$, channel scores $\{\sigma_i^{(\ell,h)}\}$, $\rho$
        \ENSURE retained $\K$ pairs $\{\mathcal{S}_k^{(\ell,h)}\}$ and $\V$ channels $\{\mathcal{S}_v^{(\ell,h)}\}$
        \FOR{$\ell=1$ to $L$}
            \STATE $\sigma_\ell\gets\mathrm{mean}_{h,p}\bigl(\sigma_p^{(\ell,h)}\bigr)$ \textit{\# layer score}
        \ENDFOR
        \STATE \textit{\# layer-adaptive pair budgets with mean retain $\rho$}
        \STATE Set $\{m_\ell\}$ with $m_\ell\propto\sigma_\ell$ and $\frac{1}{L}\sum_{\ell}m_\ell/|\ropepair|{=}\rho$
        \FOR{$\ell=1$ to $L$}
            \FOR{$h=1$ to $H_{kv}$}
                \STATE $\mathcal{S}_k^{(\ell,h)}\gets$ the $m_\ell$ pairs with largest $\sigma_p^{(\ell,h)}$
                \STATE $\mathcal{S}_v^{(\ell,h)}\gets$ the $2m_\ell$ channels with largest $\sigma_i^{(\ell,h)}$
            \ENDFOR
        \ENDFOR
        \STATE \textbf{return} $\{\mathcal{S}_k^{(\ell,h)},\mathcal{S}_v^{(\ell,h)}\}$
    \end{algorithmic}
\end{algorithm}

\subsection{Step 3: RAP Construction}
Once each head retains $m_\ell$ pairs, we form the pruned matrix $\widetilde\W_k\in\mathbb{R}^{\hat{D}\times 2m_\ell}$ from those columns and a binary expansion $\E_k\in\mathbb{R}^{2m_\ell\times D}$ that maps back to the original positions:
\begin{equation}
    (\E_k)_{i,j}=\begin{cases}1 & \text{if }\text{RAPIndex}[i]=j,\\ 0 & \text{otherwise.}\end{cases}
    \label{eq:pruned-decomposition}
\end{equation}
Because $\E_k$ is pair-preserving, it commutes with RoPE, so
\begin{align}
    \Q\K^\top &\approx \rope(\X\W_q\E_k^\top)\,\rope(\X\widetilde\W_k)^\top \nonumber\\
              &= \rope(\X\widetilde\W_q)\,\rope(\X\widetilde\W_k)^\top,
    \label{eq:latent-score}
\end{align}
with $\widetilde\W_q=\W_q\E_k^\top\in\mathbb{R}^{\hat D\times2m_\ell}$ fused into $\W_q$.
The cache stores only $\K_{\text{latent}}=\rope(\X\widetilde\W_k)\in\mathbb{R}^{S\times2m_\ell}$.
In practice, $\widetilde\W_q$ and $\widetilde\W_k$ are constructed by index selection rather than matrix multiplication, so the offline cost is negligible.
Unlike low-rank decomposition, nothing needs to be reconstructed online, so RAP adds no runtime reconstruction overhead.
We keep the original softmax scale $1/\sqrt{D}$ (not $1/\sqrt{2m_\ell}$), so removing a pair is exactly equivalent to zeroing it in the full dimension, an identity rather than a tuned choice (derived in \Cref{app:method}) and the basis for \Cref{sec:rap-theory}.

\subsection{Step 4: Accuracy Recovery}
Structured pruning is typically followed by a lightweight recovery step~\citep{llm-pruner}. We use knowledge distillation (KD)~\citep{hinton2015distilling}.
We align the compressed student to the dense teacher with a cheap Low-Rank Adaptation (LoRA)~\citep{lora} update, $\W' = \W + \mathbf{L}_{\text{down}}\mathbf{L}_{\text{up}}$ with rank $r{=}8$ ($<1\%$ of model size), applied to $\W_q,\W_k,\W_v,\W_o$ and merged before deployment.
The loss combines cross-entropy with KL distillation, $\mathcal{L}=\mathcal{L}_{\text{CE}}+\mathcal{L}_{\text{KD}}$, where $\mathcal{L}_{\text{KD}}=\mathbb{E}_{x\sim\mathcal{D}}[\mathrm{KL}(P_{\text{student}}\|P_{\text{teacher}})]$.
Without recovery, RAP already supports about $10\%$ \KV{} compression with usable quality.
With this LoRA--KD recipe it reaches $30\%$ (\Cref{sec:experiments}).

\subsection{Practical Considerations}
\label{sec:rap-serving}
\textbf{RoPE variants.}
The same pair-preserving pruning is compatible with RoPE variants such as YaRN~\citep{yarn} and Circle-RoPE~\citep{circle-rope}.
They change scaling, frequency, or pairing layout, but keep the $2{\times}2$ rotation blocks that RAP assumes.

\noindent \textbf{Hybrid compression.}
RAP applies only where RoPE is used.
In standard LLMs this is the $\Q/\K$ path, so we run RAP on $\W_k$ and channel-prune $\W_v$ to $2m_\ell$ channels, then fuse into $\W_q$ and $\W_o$ to raise accuracy while keeping decode reconstruction-free.
For partial-RoPE designs such as iRoPE, the system applies RAP only to RoPE-enabled layers while applying channel pruning to others (\Cref{app:rope_adoption}).

\noindent \textbf{Non-contiguous RoPE.}
After RAP, different heads may retain different pairs, so RoPE must index non-contiguous $\cos/\sin$ entries.
Naive PyTorch indexing materializes extra copies, which introduce overhead.
So we reimplement this step with a Triton kernel that reads the required entries directly, removing the overhead (\Cref{app:efficiency}).

\section{Theoretical Analysis of RAP Pair Selection}
\label{sec:rap-theory}
RAP selects existing RoPE coordinate pairs rather than projecting them into a learned basis, so for a fixed retained set its approximation error comes only from the removed pairs.
We give a standard second-order estimate of this loss increase and state its assumptions.
Consider one attention head, omit layer and head indices, and write $m$ for the budget $m_\ell$ of \Cref{sec:allocation}.
RAP partitions the columns of $\W_k\in\mathbb{R}^{\hat D\times D}$ into RoPE pairs $\mathcal{P}$ and retains $\mathcal{S}\subset\mathcal{P}$ with $|\mathcal{S}|=m$.
Let $\E_k\in\mathbb{R}^{2m\times D}$ be the corresponding selection matrix and define the full-dimensional mask $\M_{\mathcal S}\triangleq\E_k^\top\E_k$.
Then $\widetilde\W_k=\W_k\E_k^\top$ is the deployed reduced projection, while $\W_k\M_{\mathcal S}$ is its full-dimensional zeroed representation.
Because $\M_{\mathcal S}$ preserves complete RoPE blocks, it commutes with the full-dimensional RoPE rotation.
Together with the original softmax scale $1/\sqrt D$, reduced RAP attention is therefore functionally equivalent to replacing $\W_k$ by $\W_k\M_{\mathcal S}$ in full-dimensional attention.

\begin{proposition}[Second-order loss estimate]
\label{prop:loss_estimate}
Let $\mathcal{L}$ denote the pre-recovery calibration loss as a function of $\W_k$, with all other parameters held fixed.
Assume (i) the linear term and higher-order remainder of its Taylor expansion are negligible for the perturbation $\Delta\W_k=\W_k(\M_{\mathcal S}-\mathbf{I}_D)$, and (ii) the Hessian is approximated by the empirical diagonal Fisher~\citep{lecun-obd, molchanov2019importance, kwon2022fast}.
Then
\begin{equation}
\Delta\mathcal{L}\triangleq\mathcal{L}(\W_k\M_{\mathcal S})-\mathcal{L}(\W_k)
\approx\frac{1}{2}\sum_{p\notin\mathcal{S}}s_p,
\end{equation}
where $s_p=\sum_{i\in p}\sum_nF_{n,i}(\W_k)_{n,i}^2$ is the weight-scaled Fisher saliency of pair $p$, and $F_{n,i}$ is the entrywise Fisher in Eq.~\eqref{eq:fisher}.
\end{proposition}
\noindent(Derivation in \Cref{app:method}.)
This is an estimate rather than a bound, as the pruning perturbation is not local.

\begin{corollary}[Greedy selection]
For a fixed per-head budget $m$, the estimate is minimized by retaining the $m$ pairs with the largest $s_p$.
\end{corollary}

\noindent The estimate motivates additive scoring and selection at RoPE-pair granularity.
Guided by this pair-level objective, RAP uses the root-Fisher aggregate $\sigma_p$ of Eq.~\eqref{eq:score} as its practical saliency score, whose effectiveness is evaluated in \Cref{sec:ablation}.
The analysis is pair-level only; cross-layer and $\K/\V$ budgets are handled separately by the allocation strategy in \Cref{sec:allocation}.
\begin{table*}[t]
    \centering
    \small
    \setlength{\tabcolsep}{3.5pt}
    \begin{tabular}{l|l l cccccc ccc}
    \hhline{------------}
    \multicolumn{1}{l}{\textbf{Model}} & & \textbf{Method} & \textbf{\KV{}} & \textbf{Attn Par.} & \textbf{Attn FLOPs} & \textbf{Prefill} & \textbf{Decode} & \textbf{Rec.} & \textbf{PPL}$\downarrow$ & \textbf{ZS}$\uparrow$ & \textbf{LB}$\uparrow$ \\
    \hhline{-|-----------}
    \multirow{5}{*}{Llama-3-8B} & & Baseline & 100\% & 100\% & 100\% & 100\% & 100\% & No & 8.28 & 0.677 & 60.0 \\
    \hhline{~|-----------}
     & \multirow{2}{*}{Pruning} & KVPruner & 70\%\,\gdn & 70.0\%\,\gdn & 70.0\%\,\gdn & 83\%\,\gdn & 77\%\,\gdn & No & 397 & 0.594 & 3.4 \\
     & & \textbf{RAP} & 70\%\,\gdn & \textbf{70.3\%}\,\gdn & \textbf{70.3\%}\,\gdn & \textbf{83\%}\,\gdn & \textbf{77\%}\,\gdn & \textbf{No} & 9.12 & 0.642 & 51.4 \\
    \hhline{~|-----------}
     & \multirow{2}{*}{Low-rank} & SVD-LLM & 70\%\,\gdn & 97.6\%\,\gdn & 97.6\%\,\gdn & 100\%\,\rup & 196\%\,\rup & Yes & 136.45 & 0.389 & 8.5 \\
     & & PaLU & 70\%\,\gdn & 82.4\%\,\gdn & 82.4\%\,\gdn & 101\%\,\rup & 112\%\,\rup & Yes & \textbf{8.59} & \textbf{0.671} & \textbf{55.3} \\
    \hhline{-|-----------}
    \multirow{5}{*}{Mistral-7B} & & Baseline & 100\% & 100\% & 100\% & 100\% & 100\% & No & 5.32 & 0.685 & 57.4 \\
    \hhline{~|-----------}
     & \multirow{2}{*}{Pruning} & KVPruner & 70\%\,\gdn & 70.0\%\,\gdn & 70.0\%\,\gdn & 83\%\,\gdn & 76\%\,\gdn & No & 78.9 & 0.629 & 2.2 \\
     & & \textbf{RAP} & 70\%\,\gdn & \textbf{70.3\%}\,\gdn & \textbf{70.3\%}\,\gdn & \textbf{83\%}\,\gdn & \textbf{76\%}\,\gdn & \textbf{No} & 5.69 & 0.650 & 50.9 \\
    \hhline{~|-----------}
     & \multirow{2}{*}{Low-rank} & SVD-LLM & 70\%\,\gdn & 97.6\%\,\gdn & 97.6\%\,\gdn & 101\%\,\rup & 200\%\,\rup & Yes & 42.26 & 0.418 & 6.4 \\
     & & PaLU & 70\%\,\gdn & 82.4\%\,\gdn & 82.4\%\,\gdn & 100\%\,\rup & 111\%\,\rup & Yes & \textbf{5.38} & \textbf{0.673} & \textbf{56.1} \\
    \hhline{-|-----------}
    \end{tabular}
    \caption{Iso-\KV{} comparison of quality and attention cost at $\rho{=}0.7$ (post-KD). Best of PaLU/RAP in \textbf{bold}.}
    \label{tab:main_results}
    \label{tab:comprehensive}
    \end{table*}
    
    \begin{figure*}[!t]
    \centering
    \includegraphics{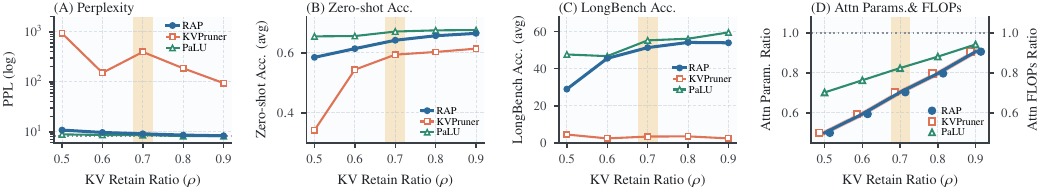}
    \caption{Iso-\KV{} comparison of accuracy and attention cost on Llama-3-8B. \protect\sandswatch{} marks the practical operating point at $\rho{=}0.7$.}
    \label{fig:accuracy_fig}
    \end{figure*}

\section{Experiments}
\label{sec:experiments}
We evaluate RAP at \textbf{iso-\KV{}-cache}: every method uses the same $\rho$ fraction of \KV{} channels (\Cref{tab:capability}).
At the practical operating point $\rho{=}0.7$, RAP maintains usable accuracy while co-scaling attention parameters and FLOPs with $\rho$ and staying reconstruction-free, which cuts Prefill/Decode latency. 

\subsection{Setup}
\label{sec:setup}
\textbf{Default model.}
We use Meta-Llama-3-8B-Instruct as the default checkpoint and refer to it as Llama-3-8B.

\noindent\textbf{Generality.}
To check that the findings are not model-specific, we also evaluate on different model types (Mistral, and Qwen) and scales (3B, 14B), plus a controlled $\theta_{\text{base}}$ sweep on fixed Llama-3-8B for the older low-base regime, and Llama-2-7B to reproduce published KVPruner recoverability (\Cref{app:baselines,app:extended}).
\Cref{app:extended} reports their perplexity and LongBench results.

\noindent\textbf{Metrics.}
We report WikiText-2~\citep{wikitext2} test PPL in fp32 over contiguous non-overlapping windows of 2048 tokens.
We also report zero-shot accuracy (ZS) averaged over six tasks (OpenBookQA, HellaSwag, PIQA, ARC-Easy, ARC-Challenge with acc\_norm, and WinoGrande with acc), and LongBench~\citep{longbench} accuracy (LB) averaged over five tasks (TriviaQA, Qasper, TREC, SAMSum, and LCC, 50 examples per task). The full LongBench matrix is in \Cref{app:extended}.
On the efficiency side we report attention parameters (Attn Par.) and attention FLOPs (Attn FLOPs) as a percentage of the dense model, Prefill and Decode as attention-module latency relative to dense, and Rec.\ as a Yes/No flag for per-step $\K$ reconstruction.

\noindent\textbf{Baselines.}
The most relevant baseline is KVPruner~\citep{kv-pruner}, the RoPE-agnostic channel-pruning counterpart of RAP, which selects individual $\K$/$\V$ channels rather than whole RoPE pairs.
For weight-side low-rank decomposition we compare SVD-LLM~\citep{svd-llm} and PaLU~\citep{palu} (official PaLU decomposition under the shared KD recipe).
Under iso-\KV{}, matched cache does not imply matched compute.
SVD-LLM and PaLU write $\W\approx\A\B$ and keep the low-rank factors online, so Attn Par.\ and FLOPs stay above $\rho$ times dense (\Cref{tab:rap_complexity,app:cost_model}), and both must reconstruct $\K$ at every decode step.
Activation-side methods such as MatryoshkaKV~\citep{matryoshkakv} leave $\W_k,\W_v$ untouched, so they do not reduce parameters or projection FLOPs, and we do not treat them as direct baselines.
Token eviction~\citep{snapkv, h2o, kvzip} shortens the token axis while RAP narrows the channel axis, so the two multiply rather than compete.
We still ran MatryoshkaKV under the shared recipe; we compose RAP with quantization in \Cref{sec:analysis}.
Baseline details, including the index-aware KVPruner steelman and non-converging methods, are in \Cref{app:baselines}.
Unless noted, RAP, KVPruner, and PaLU use the same Fisher allocation and LoRA--KD recovery (\Cref{alg:allocation}).
We sweep $\rho\in\{0.5,0.6,0.7,0.8,0.9\}$ and take $\rho{=}0.7$ as the practical point.

\noindent\textbf{Calibration and recovery.}
We estimate Fisher scores on 32 sequences of 1024 tokens from WikiText-2 train (\Cref{alg:rap}).
We conduct accuracy recovery with LoRA--KD ($r{=}8$, $\alpha{=}16$, dropout 0.05, temperature 2.0, CE+KD) on the same corpus, with a C4-en control in \Cref{app:repro} ruling out domain overlap.

\noindent All runs are on one NVIDIA A100 80GB (PyTorch 2.4.0+cu121, transformers 4.57). See \Cref{app:repro} for full reproducibility details and \Cref{app:limitations} for limitations.

\subsection{Main Results}
\label{sec:main_results}
\label{sec:accuracy}
\label{sec:vs_kv-pruner}
\label{sec:efficiency}
\label{sec:vs_palu}
\Cref{tab:main_results} reports the iso-\KV{} snapshot at the practical operating point $\rho{=}0.7$ on Llama-3-8B and Mistral-7B.
Every method keeps the same fraction of dense \KV{} channels, so differences come from \emph{how} that budget is spent: channel vs.\ RoPE-pair pruning, or low-rank factorization with reconstruction.
RAP is the only row that stays close to baseline on PPL/ZS/LB, sits near $70\%$ Attn Par./FLOPs, and reports Prefill/Decode below dense \emph{without} Rec.
KVPruner matches RAP on cost and latency but collapses on quality. PaLU is slightly stronger on quality but pays higher Attn cost and a reconstruction tax on decode.

\noindent\textbf{Quality.}
\Cref{fig:accuracy_fig}(A)(B)(C) sweeps $\rho{=}0.5$--$0.9$ on Llama-3-8B.
KVPruner is the matched-cost counterpart of RAP: at every $\rho$ it keeps the same Attn Par./FLOPs, so the only difference is the prune unit (individual channels vs.\ whole RoPE pairs).
Across the sweep, KVPruner stays $11\text{--}86\times$ worse on Llama PPL and collapses on LongBench ($2.4$--$4.5$), while RAP remains in a usable band.
Index bookkeeping is not the missing piece: an index-aware KVPruner that rotates each surviving channel at its original frequency still trails RAP by $10\text{--}84\times$ post-KD, because its half-pairs remain orphaned (\Cref{app:baselines}).
At $\rho{=}0.7$, \Cref{tab:main_results} shows RAP close to PaLU and far above KVPruner on both models; KVPruner can still look competitive on zero-shot, but it fails on LongBench.
ZS is short-prompt multiple choice and only weakly sensitive to positional damage, whereas long-context evaluation is far more sensitive to compression~\citep{mekala-etal-2025-quantization, li2025scbench} and to the broken pairs we diagnose in \Cref{sec:analysis}.
PaLU holds better accuracy under iso-\KV{} only with extra Attn parameters: once parameters are matched (RAP at $\rho{=}0.7$ vs.\ PaLU at $\rho{=}0.5$), RAP keeps $0.70\times$ cache vs.\ PaLU $0.52\times$ at matched ${\approx}70\%$ Attn, and leads on LongBench ($51.4$ vs.\ $47.7$ on Llama, \Cref{app:efficiency}).

\noindent\textbf{Attention cost.}
Matched \KV{} does not imply matched compute.
Low-rank methods write $\W\approx\A\B$, which under iso-\KV{} would roughly double the Attn parameters of RAP if both factors stayed online.
PaLU fuses $\B$ into the following matrix when possible, but RoPE blocks that fusion on $\K$, so a $\K$-side factor remains online as an $O(\rho/H)$ overhead (\Cref{tab:rap_complexity}).
At iso-\KV{} $\rho{=}0.7$ the cache matches by construction, but that leftover factor inflates Attn Par./FLOPs to ${\approx}82\%$ while RAP stays at ${\approx}70\%$ (\Cref{tab:main_results}).
\Cref{fig:accuracy_fig}(D) plots the empirical ratios versus $\rho$: RAP sits on the ideal $\rho$ line, while PaLU and SVD-LLM stay strictly above it.
To match RAP's ${\approx}70\%$ Attn budget at $\rho{=}0.7$, PaLU must drop to $\rho{=}0.5$, which is the matched-compute setting in \Cref{app:efficiency}.

\begin{figure*}[!t]
\centering
\begin{minipage}[t]{0.495\textwidth}
\centering
\includegraphics[width=0.495\linewidth]{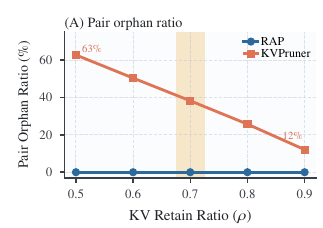}%
\includegraphics[width=0.495\linewidth]{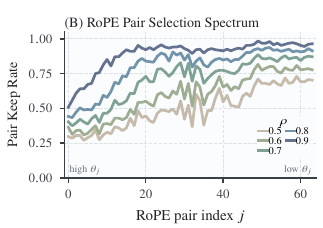}
\captionof{figure}{Pair integrity study (Llama-3-8B).}
\label{fig:kept_pairs}
\label{fig:kept_wavelength}
\end{minipage}\hfill
\begin{minipage}[t]{0.495\textwidth}
\centering
\includegraphics[width=0.495\linewidth]{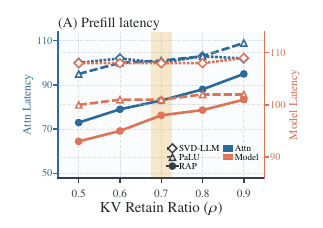}%
\includegraphics[width=0.495\linewidth]{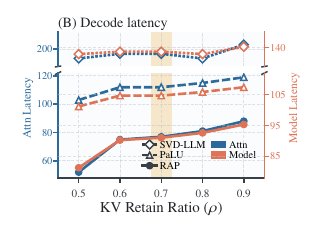}
\captionof{figure}{Prefill/Decode efficiency (Llama-3-8B).}
\label{fig:efficiency}
\label{fig:flops}
\label{fig:decode}
\end{minipage}
\end{figure*}

\begin{figure*}[!t]
\centering
\begin{minipage}[t]{0.32\textwidth}
\centering
\includegraphics[width=\linewidth]{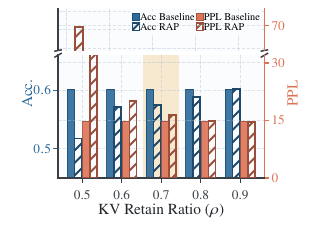}
\captionof{figure}{RAP with 4-bit quantization.}
\label{fig:quant_summary}
\end{minipage}\hfill
\begin{minipage}[t]{0.32\textwidth}
\centering
\includegraphics[width=\linewidth]{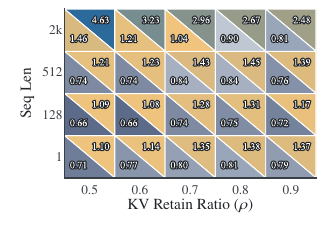}
\captionof{figure}{RoPE kernel speedup.}
\label{fig:rope_kernel_heatmap}
\end{minipage}\hfill
\begin{minipage}[t]{0.32\textwidth}
\centering
\includegraphics[width=\linewidth]{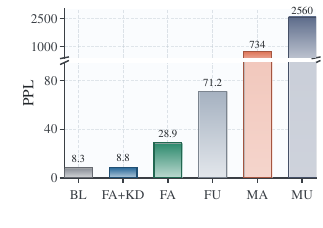}
\captionof{figure}{Ablation study summary.}
\label{fig:ablation_allocation}
\end{minipage}
\end{figure*}
\begin{table}[t]
    \centering
    \small
    \setlength{\tabcolsep}{4pt}
    \begin{tabular}{lccc}
    \toprule
    \textbf{Method} & \textbf{\KV{}-Cache} & \textbf{Parameters} & \textbf{FLOPs} \\
    \midrule
    Baseline ($\mathcal{B}$) & $2SD$ & $2HD^2$ & $4SHD^2$ \\
    SVD-LLM & $\rho\mathcal{B}$ & $(\rho+\frac{\rho}{H})\mathcal{B}$ & $(\rho+\frac{\rho}{H})\mathcal{B}$ \\
    PaLU & $\rho\mathcal{B}$ & $(\rho+\frac{\rho}{2H})\mathcal{B}$ & $(\rho+\frac{\rho}{2H})\mathcal{B}$ \\
    \textbf{RAP} & $\boldsymbol{\rho\mathcal{B}}$ & $\boldsymbol{\rho\mathcal{B}}$ & $\boldsymbol{\rho\mathcal{B}}$ \\
    \bottomrule
    \end{tabular}
    \caption{Attention cost scaling with $\rho$.}
    \label{tab:rap_complexity}
\end{table}
\noindent\textbf{Latency.}
\Cref{fig:efficiency}(A)(B) plots attention-module Prefill/Decode latency versus $\rho$, reported as a percentage of the dense baseline (lower is better).
At $\rho{=}0.7$, RAP cuts attention Prefill/Decode to $83\%$/$77\%$ of dense, whereas SVD-LLM and PaLU stay at or above baseline on prefill and well above on decode.
The gains survive end to end: full-model decode reaches $1.05$--$1.11\times$ dense at 16k--32k and peak memory stays below dense at every batch we could run, while PaLU runs out at batch~4 (\Cref{app:efficiency}).

\subsection{Analysis}
\label{sec:analysis}
\label{sec:ablation}
The main results show that RAP preserves accuracy at $\rho{=}0.7$ while cutting Attn parameters/FLOPs and Prefill/Decode latency.
Here we conduct further analysis of the algorithm.
We first explain why Fisher-based pair selection is effective, then run design ablations over scoring and allocation, check composition with 4-bit weights, and evaluate the fused RoPE kernel for non-contiguous indexing.

\noindent\textbf{Pair integrity.}
We first ask why the Fisher pair unit works.
\Cref{fig:kept_pairs}(A) measures how often channel pruning splits a RoPE pair.
The \emph{pair orphan ratio} is the fraction of kept $\K$ channels whose RoPE partner was pruned.
As compression grows ($\rho{=}0.9{\to}0.5$), KVPruner's orphan ratio rises from $12.1\%$ to $62.7\%$. Equivalently, it breaks $21.5\%$--$77.1\%$ of the RoPE pairs it touches.
RAP stays at $0\%$ by construction: every decision keeps or drops a whole pair.
That structural gap explains the matched-cost collapse in \Cref{sec:accuracy}.
\Cref{fig:kept_pairs}(B) asks which pairs RAP keeps.
In Eq.~\eqref{eq:rope}, pair $j$ rotates at $\theta_j=\theta_{\text{base}}^{-2j/D}$.
Small $j$ has large $\theta_j$ to encode short-range position. Large $j$ has small $\theta_j$, rotates slowly, and carries longer-range position~\citep{rope-dim-inefficiency,triattention}.
\Cref{fig:kept_pairs}(B) keeps rate against $j$: Fisher keeps the large-$j$ (long-wavelength) pairs more often than the small-$j$ ones.
Beyond being RoPE-safe, the scorer spends more budget on pairs that matter more for long-range position.

\noindent\textbf{Design ablations.}
We next ablate the design choices to validate RAP's recipe.
We compare three factors at $\rho{=}0.7$ under a pooled $\K$/$\V$ budget (\Cref{fig:ablation_allocation}): scoring (Fisher vs.\ magnitude), layer budget (adaptive vs.\ uniform), and LoRA--KD.
In the figure, the first letter is the scorer (F$=$Fisher, M$=$Magnitude) and the second is the layer policy (A$=$Adaptive, U$=$Uniform). Unless marked $+{\rm KD}$, the bar is prune-only.
MU is the naive baseline: magnitude scores with a flat per-layer retain ratio, whose PPL is unusable.
Replacing magnitude by Fisher, or uniform by adaptive, each helps on its own. Combining both gives FA (Fisher$+$Adaptive), the best prune-only setting.
Applying LoRA--KD to FA (FA$+$KD) then brings PPL back to near-baseline.
We also sweep the LoRA rank over $r{=}4/8/16$ and obtain PPL $9.52/9.35/9.37$, so recovery does not hinge on a tuned rank.
A further design choice is how to split the channel budget between $\K$ and $\V$ (\Cref{sec:allocation}).
Pooling the two into one Fisher allocation severely under-allocates $\K$ and costs $8$ LongBench points at $\rho{=}0.7$ (\Cref{app:method}), so RAP gives $\K$ and $\V$ the same budget instead.
We also conduct a split experiment that assigns $0.35/0.50/0.65$ of the channels to $\K$. PPL is worse at both extremes and best at $0.50$, so the equal split is not an arbitrary pick.
Details on KD, $\K$/$\V$ split, and RoPE-base studies are in \Cref{app:extended}.

\noindent\textbf{4-bit weights.}
Finally, we check that RAP survives 4-bit weight quantization on top of pruning.
\Cref{fig:quant_summary} compares RAP against a 4-bit dense baseline across $\rho$.
At $\rho{=}0.7$ RAP holds $0.575$ accuracy against $0.602$ dense and $16.5$ PPL against $14.8$, which again places the usable point at $\rho{=}0.7$.

\noindent\textbf{RoPE kernel.}
After RAP, the kept $\K$ channels are no longer contiguous inside a head, which can introduce RoPE overhead at inference.
We find that this overhead is an artifact of the framework implementation, not a fundamental cost of RAP.
In PyTorch's default RoPE, non-contiguous indexing allocates a new memory buffer before applying $\cos/\sin$, which is the root cause of the overhead.
We reimplement RoPE in Triton so that non-contiguous access bypasses the extra buffer copy.
\Cref{fig:rope_kernel_heatmap} plots speedup versus dense RoPE over sequence length and $\rho$.
Each cell splits into two triangles: lower-left is Torch and upper-right is Triton.
Torch often falls below $1\times$ from the copy, while Triton reaches $4.6\times$ at long $S$.
RoPE is under $1\%$ of end-to-end time, so the Prefill/Decode speedup mainly comes from dropping reconstruction and shrinking parameters, not from this kernel.

\section{Conclusion}
We introduce RAP, a structured pruning method that removes RoPE-aligned column pairs under sensitivity-guided budgets.
The pair-preserving selection commutes with RoPE, so it fuses into $\W_q$ offline and decode needs no reconstruction.
RAP is a drop-in method that shrinks the \KV{} cache, parameters, and FLOPs together on RoPE-based LLMs and stays near the strongest low-rank baseline at $\rho{=}0.7$.

% Separate the 7-page body from references + appendix (submission only;
% remove for AAAI camera-ready, which disallows page breaks).
\clearpage

\bibliography{rap}

@inproceedings{palu,
  title={Palu: KV-cache compression with low-rank projection},
  author={Chang, Chi-Chih and Lin, Wei-Cheng and Lin, Chien-Yu and Chen, Chong-Yan and Hu, Yu-Fang and Wang, Pei-Shuo and Huang, Ning-Chi and Ceze, Luis and Abdelfattah, Mohamed and Wu, Kai-Chiang},
  booktitle={International Conference on Learning Representations},
  volume={2025},
  pages={50222--50249},
  year={2025}
}

@inproceedings{sparsegpt,
  title={Sparsegpt: Massive language models can be accurately pruned in one-shot},
  author={Frantar, Elias and Alistarh, Dan},
  booktitle={International conference on machine learning},
  pages={10323--10337},
  year={2023},
  organization={PMLR}
}

@inproceedings{svd-llm,
title={{SVD}-{LLM}: Truncation-aware Singular Value Decomposition for Large Language Model Compression},
author={Xin Wang and Yu Zheng and Zhongwei Wan and Mi Zhang},
booktitle={The Thirteenth International Conference on Learning Representations},
year={2025},
url={https://openreview.net/forum?id=LNYIUouhdt}
}

@article{llm-pruner,
  title={Llm-pruner: On the structural pruning of large language models},
  author={Ma, Xinyin and Fang, Gongfan and Wang, Xinchao},
  journal={Advances in neural information processing systems},
  volume={36},
  pages={21702--21720},
  year={2023}
}

@inproceedings{wanda,
  title={A simple and effective pruning approach for large language models},
  author={Sun, Mingjie and Liu, Zhuang and Bair, Anna and Kolter, Zico},
  booktitle={International Conference on Learning Representations},
  volume={2024},
  pages={4942--4964},
  year={2024}
}

@article{snapkv,
  title={Snapkv: Llm knows what you are looking for before generation},
  author={Li, Yuhong and Huang, Yingbing and Yang, Bowen and Venkitesh, Bharat and Locatelli, Acyr and Ye, Hanchen and Cai, Tianle and Lewis, Patrick and Chen, Deming},
  journal={Advances in Neural Information Processing Systems},
  volume={37},
  pages={22947--22970},
  year={2024}
}

@article{longbench,
  title={Longbench: A bilingual, multitask benchmark for long context understanding},
  author={Bai, Yushi and Lv, Xin and Zhang, Jiajie and Lyu, Hongchang and Tang, Jiankai and Huang, Zhidian and Du, Zhengxiao and Liu, Xiao and Zeng, Aohan and Hou, Lei and others},
  journal={arXiv preprint arXiv:2308.14508},
  year={2023}
}

@article{kvcachechallenge,
  title={Challenges in deploying long-context transformers: A theoretical peak performance analysis},
  author={Fu, Yao},
  journal={arXiv preprint arXiv:2405.08944},
  year={2024}
}

@article{rope,
author = {Su, Jianlin and Ahmed, Murtadha and Lu, Yu and Pan, Shengfeng and Bo, Wen and Liu, Yunfeng},
title = {RoFormer: Enhanced transformer with Rotary Position Embedding},
year = {2024},
issue_date = {Feb 2024},
publisher = {Elsevier Science Publishers B. V.},
address = {NLD},
volume = {568},
number = {C},
issn = {0925-2312},
url = {https://doi.org/10.1016/j.neucom.2023.127063},
doi = {10.1016/j.neucom.2023.127063},
journal = {Neurocomput.},
month = feb,
numpages = {12}
}

@inproceedings{lora,
title={Lo{RA}: Low-Rank Adaptation of Large Language Models},
author={Edward J Hu and yelong shen and Phillip Wallis and Zeyuan Allen-Zhu and Yuanzhi Li and Shean Wang and Lu Wang and Weizhu Chen},
booktitle={International Conference on Learning Representations},
year={2022},
url={https://openreview.net/forum?id=nZeVKeeFYf9}
}

@article{attention,
  title={Attention is all you need},
  author={Vaswani, Ashish and Shazeer, Noam and Parmar, Niki and Uszkoreit, Jakob and Jones, Llion and Gomez, Aidan N and Kaiser, {\L}ukasz and Polosukhin, Illia},
  journal={Advances in neural information processing systems},
  volume={30},
  year={2017}
}

@inproceedings{kivi,
author = {Liu, Zirui and Yuan, Jiayi and Jin, Hongye and Zhong, Shaochen (Henry) and Xu, Zhaozhuo and Braverman, Vladimir and Chen, Beidi and Hu, Xia},
title = {KIVI: a tuning-free asymmetric 2bit quantization for KV cache},
year = {2024},
publisher = {JMLR.org},
booktitle = {Proceedings of the 41st International Conference on Machine Learning},
articleno = {1311},
numpages = {13},
location = {Vienna, Austria},
series = {ICML'24}
}

@article{kvquant,
  title={Kvquant: Towards 10 million context length llm inference with kv cache quantization},
  author={Hooper, Coleman and Kim, Sehoon and Mohammadzadeh, Hiva and Mahoney, Michael W and Shao, Yakun S and Keutzer, Kurt and Gholami, Amir},
  journal={Advances in Neural Information Processing Systems},
  volume={37},
  pages={1270--1303},
  year={2024}
}

@article{h2o,
  title={H2o: Heavy-hitter oracle for efficient generative inference of large language models},
  author={Zhang, Zhenyu and Sheng, Ying and Zhou, Tianyi and Chen, Tianlong and Zheng, Lianmin and Cai, Ruisi and Song, Zhao and Tian, Yuandong and R{\'e}, Christopher and Barrett, Clark and others},
  journal={Advances in Neural Information Processing Systems},
  volume={36},
  pages={34661--34710},
  year={2023}
}

@article{hinton2015distilling,
  title={Distilling the knowledge in a neural network},
  author={Hinton, Geoffrey and Vinyals, Oriol and Dean, Jeff},
  journal={arXiv preprint arXiv:1503.02531},
  year={2015}
}

@article{achiam2023gpt,
  title={Gpt-4 technical report},
  author={Achiam, Josh and Adler, Steven and Agarwal, Sandhini and Ahmad, Lama and Akkaya, Ilge and Aleman, Florencia Leoni and Almeida, Diogo and Altenschmidt, Janko and Altman, Sam and Anadkat, Shyamal and others},
  journal={arXiv preprint arXiv:2303.08774},
  year={2023}
}

@article{grattafiori2024llama,
  title={The llama 3 herd of models},
  author={Grattafiori, Aaron and Dubey, Abhimanyu and Jauhri, Abhinav and Pandey, Abhinav and Kadian, Abhishek and Al-Dahle, Ahmad and Letman, Aiesha and Mathur, Akhil and Schelten, Alan and Vaughan, Alex and others},
  journal={arXiv preprint arXiv:2407.21783},
  year={2024}
}

@article{bai2023qwen,
  title={Qwen technical report},
  author={Bai, Jinze and Bai, Shuai and Chu, Yunfei and Cui, Zeyu and Dang, Kai and Deng, Xiaodong and Fan, Yang and Ge, Wenbin and Han, Yu and Huang, Fei and others},
  journal={arXiv preprint arXiv:2309.16609},
  year={2023}
}

@article{liu2024deepseek,
  title={Deepseek-v3 technical report},
  author={Liu, Aixin and Feng, Bei and Xue, Bing and Wang, Bingxuan and Wu, Bochao and Lu, Chengda and Zhao, Chenggang and Deng, Chengqi and Zhang, Chenyu and Ruan, Chong and others},
  journal={arXiv preprint arXiv:2412.19437},
  year={2024}
}

@article{gholami2024ai,
  title={Ai and memory wall},
  author={Gholami, Amir and Yao, Zhewei and Kim, Sehoon and Hooper, Coleman and Mahoney, Michael W and Keutzer, Kurt},
  journal={IEEE Micro},
  volume={44},
  number={3},
  pages={33--39},
  year={2024},
  publisher={IEEE}
}

@misc{llama4,
  title = {{Llama-4-Scout-17B-16E-Instruct}},
  author = {{Meta AI}},
  year = {2025},
  howpublished = {\url{https://huggingface.co/meta-llama/Llama-4-Scout-17B-16E-Instruct}},
  note = {Model card}
}

@article{farabet2025introducing,
  title = {Introducing Gemma 3: The most capable model you can run on a single GPU or TPU},
  author = {Farabet, Clement and Warkentin, Tris},
  journal = {Google Blogs},
  year = {2025}
}

@misc{gptoss,
  title = {{gpt-oss-20b}},
  author = {{OpenAI}},
  year = {2025},
  howpublished = {\url{https://huggingface.co/openai/gpt-oss-20b}},
  note = {Model card}
}

@misc{mistral32,
  title = {{Mistral-Small-3.2-24B-Instruct-2506}},
  author = {{Mistral AI}},
  year = {2025},
  howpublished = {\url{https://huggingface.co/mistralai/Mistral-Small-3.2-24B-Instruct-2506}},
  note = {Model card}
}

@misc{qwen3next,
  title = {{Qwen3-Next-80B-A3B-Instruct}},
  author = {{Qwen Team}},
  year = {2025},
  howpublished = {\url{https://huggingface.co/Qwen/Qwen3-Next-80B-A3B-Instruct}},
  note = {Model card}
}

@misc{glm47,
  title = {{GLM-4.7}},
  author = {{Zhipu AI}},
  year = {2025},
  howpublished = {\url{https://huggingface.co/zai-org/GLM-4.7}},
  note = {Model card}
}

@inproceedings{mekala-etal-2025-quantization,
  title={Does quantization affect models’ performance on long-context tasks?},
  author={Mekala, Anmol and Atmakuru, Anirudh and Song, Yixiao and Karpinska, Marzena and Iyyer, Mohit},
  booktitle={Proceedings of the 2025 Conference on Empirical Methods in Natural Language Processing},
  pages={9433--9481},
  year={2025}
}

@inproceedings{li2025scbench,
  title={Scbench: A kv cache-centric analysis of long-context methods},
  author={Li, Yucheng and Jiang, Huiqiang and Wu, Qianhui and Luo, Xufang and Ahn, Surin and Zhang, Chengruidong and Abdi, Amir and Li, Dongsheng and Gao, Jianfeng and Yang, Yuqing and others},
  booktitle={International Conference on Learning Representations},
  volume={2025},
  pages={66063--66093},
  year={2025}
}

@inproceedings{yarn,
title={Ya{RN}: Efficient Context Window Extension of Large Language Models},
author={Bowen Peng and Jeffrey Quesnelle and Honglu Fan and Enrico Shippole},
booktitle={The Twelfth International Conference on Learning Representations},
year={2024},
url={https://openreview.net/forum?id=wHBfxhZu1u}
}

@article{circle-rope,
  title={Circle-rope: Cone-like decoupled rotary positional embedding for large vision-language models},
  author={Wang, Chengcheng and Guo, Jianyuan and Li, Hongguang and Tian, Yuchuan and Nie, Ying and Xu, Chang and Han, Kai},
  journal={arXiv preprint arXiv:2505.16416},
  year={2025}
}

@INPROCEEDINGS{kv-pruner,
  author={Lv, Bo and Zhou, Quan and Ding, Xuanang and Wang, Yan and Ma, Zeming},
  booktitle={ICASSP 2025 - 2025 IEEE International Conference on Acoustics, Speech and Signal Processing (ICASSP)}, 
  title={KVPruner: Structural Pruning for Faster and Memory-Efficient Large Language Models}, 
  year={2025},
  volume={},
  number={},
  pages={1-5},
  doi={10.1109/ICASSP49660.2025.10889000}}

@article{elitekv,
  title={EliteKV: Scalable KV Cache Compression via RoPE Frequency Selection and Joint Low-Rank Projection},
  author={Yuhao Zhou and Sirui Song and Boyang Liu and Zhiheng Xi and Senjie Jin and Xiaoran Fan and Zhihao Zhang and Wei Li and Xuanjing Huang},
  journal={ArXiv},
  year={2025},
  volume={abs/2503.01586},
  url={https://api.semanticscholar.org/CorpusID:276776349}
}

@inproceedings{matryoshkakv,
  title={Matryoshkakv: Adaptive kv compression via trainable orthogonal projection},
  author={Lin, Bokai and Zeng, Zihao and Xiao, Zipeng and Kou, Siqi and Hou, Tianqi and Gao, Xiaofeng and Zhang, Hao and Deng, Zhijie},
  booktitle={International Conference on Learning Representations},
  volume={2025},
  pages={86669--86690},
  year={2025}
}

@article{lecun-obd,
  title={Optimal brain damage},
  author={LeCun, Yann and Denker, John and Solla, Sara},
  journal={Advances in neural information processing systems},
  volume={2},
  year={1989}
}

@inproceedings{molchanov2019importance,
  title={Importance estimation for neural network pruning},
  author={Molchanov, Pavlo and Mallya, Arun and Tyree, Stephen and Frosio, Iuri and Kautz, Jan},
  booktitle={Proceedings of the IEEE/CVF conference on computer vision and pattern recognition},
  pages={11264--11272},
  year={2019}
}

@article{kwon2022fast,
  title={A fast post-training pruning framework for transformers},
  author={Kwon, Woosuk and Kim, Sehoon and Mahoney, Michael W and Hassoun, Joseph and Keutzer, Kurt and Gholami, Amir},
  journal={Advances in Neural Information Processing Systems},
  volume={35},
  pages={24101--24116},
  year={2022}
}

@inproceedings{think,
  title={Think: Thinner key cache by query-driven pruning},
  author={Xu, Yuhui and Jie, Zhanming and Dong, Hanze and Wang, Lei and Lu, Xudong and Zhou, Aojun and Saha, Amrita and Xiong, Caiming and Sahoo, Doyen},
  booktitle={International Conference on Learning Representations},
  volume={2025},
  pages={56691--56709},
  year={2025}
}

@inproceedings{leank2025,
  title={LeanK: Learnable K Cache Channel Pruning for Efficient Decoding},
  author={Zhang, Yike and He, Zhiyuan and Jiang, Huiqiang and Zhang, Chengruidong and Yang, Yuqing and Wang, Jianyong and Qiu, Lili},
  booktitle={Proceedings of the 2025 Conference on Empirical Methods in Natural Language Processing},
  pages={31110--31125},
  year={2025}
}

@inproceedings{spark2025,
  title={Spark: Query-aware unstructured sparsity with recoverable kv cache channel pruning},
  author={Liao, Huanxuan and Xu, Yixing and He, Shizhu and Li, Guanchen and Yin, Xuanwu and Li, Dong and Barsoum, Emad and Zhao, Jun and Liu, Kang},
  booktitle={Proceedings of the AAAI Conference on Artificial Intelligence},
  volume={40},
  number={38},
  pages={31961--31969},
  year={2026}
}

@article{rotatek2026,
  title={Rotation-Aligned Key Channel Pruning for Efficient Vision-Language Model Inference},
  author={Kang, Beomseok and Jo, Dongwon and Song, Jiwon and Son, Donghwee and Kim, Jae-Joon},
  journal={arXiv preprint arXiv:2605.19218},
  year={2026}
}

@article{recalkv2025,
  title={ReCalKV: Low-Rank KV Cache Compression via Head Reordering and Offline Calibration},
  author={Yan, Xianglong and Li, Zhiteng and Zhang, Tianao and Qin, Haotong and Kong, Linghe and Zhang, Yulun and Yang, Xiaokang},
  journal={arXiv preprint arXiv:2505.24357},
  year={2025}
}

@article{wikitext2,
  title={Pointer sentinel mixture models},
  author={Merity, Stephen and Xiong, Caiming and Bradbury, James and Socher, Richard},
  journal={arXiv preprint arXiv:1609.07843},
  year={2016}
}

@inproceedings{openbookqa,
  title={Can a suit of armor conduct electricity? A new dataset for open book question answering},
  author={Mihaylov, Todor and Clark, Peter and Khot, Tushar and Sabharwal, Ashish},
  booktitle={Proceedings of the Conference on Empirical Methods in Natural Language Processing},
  year={2018}
}

@inproceedings{hellaswag,
  title={{HellaSwag}: Can a machine really finish your sentence?},
  author={Zellers, Rowan and Holtzman, Ari and Bisk, Yonatan and Farhadi, Ali and Choi, Yejin},
  booktitle={Proceedings of the Annual Meeting of the Association for Computational Linguistics},
  year={2019}
}

@inproceedings{piqa,
  title={{PIQA}: Reasoning about physical commonsense in natural language},
  author={Bisk, Yonatan and Zellers, Rowan and Gao, Jianfeng and Choi, Yejin},
  booktitle={Proceedings of the AAAI Conference on Artificial Intelligence},
  year={2020}
}

@article{arc,
  title={Think you have solved question answering? Try {ARC}, the {AI2} reasoning challenge},
  author={Clark, Peter and Cowhey, Isaac and Etzioni, Oren and Khot, Tushar and Sabharwal, Ashish and Schoenick, Carissa and Tafjord, Oyvind},
  journal={arXiv preprint arXiv:1803.05457},
  year={2018}
}

@inproceedings{winogrande,
  title={{WinoGrande}: An adversarial Winograd schema challenge at scale},
  author={Sakaguchi, Keisuke and Bras, Ronan Le and Bhagavatula, Chandra and Choi, Yejin},
  booktitle={Proceedings of the AAAI Conference on Artificial Intelligence},
  year={2020}
}

@article{c4,
  title={Exploring the limits of transfer learning with a unified text-to-text transformer},
  author={Raffel, Colin and Shazeer, Noam and Roberts, Adam and Lee, Katherine and Narang, Sharan and Matena, Michael and Zhou, Yanqi and Li, Wei and Liu, Peter J},
  journal={Journal of Machine Learning Research},
  volume={21},
  number={140},
  year={2020}
}

@inproceedings{rope-dim-inefficiency,
  title={The Rotary Position Embedding May Cause Dimension Inefficiency in Attention Heads for Long-Distance Retrieval},
  author={Chiang, Ting-Rui and Yogatama, Dani},
  booktitle={Findings of the Association for Computational Linguistics: ACL 2025},
  pages={13552--13562},
  year={2025}
}

@article{thin-keys,
  title={Thin Keys, Full Values: Reducing KV Cache via Low-Dimensional Attention Selection},
  author={Yao, Hengshuai and Chen, Xing and Murtadha, Ahmed and Wang, Guan},
  journal={arXiv preprint arXiv:2603.04427},
  year={2026}
}

@inproceedings{lethe,
  title={Lethe: Layer-and time-adaptive kv cache pruning for reasoning-intensive llm serving},
  author={Zeng, Hui and Zhao, Daming and Yang, Pengfei and Hou, WenXuan and Zheng, Tianyang and Li, Hui and Ji, Weiye and Zhai, Jidong},
  booktitle={Proceedings of the AAAI Conference on Artificial Intelligence},
  volume={40},
  number={33},
  pages={28103--28112},
  year={2026}
}

@article{triattention,
  title={TriAttention: Efficient long reasoning with trigonometric KV compression},
  author={Mao, Weian and Lin, Xi and Huang, Wei and Xie, Yuxin and Fu, Tianfu and Zhuang, Bohan and Han, Song and Chen, Yukang},
  journal={arXiv preprint arXiv:2604.04921},
  year={2026}
}

@article{kvzip,
  title={Kvzip: Query-agnostic kv cache compression with context reconstruction},
  author={Kim, Jang-Hyun and Kim, Jinuk and Kwon, Sangwoo and Lee, Jae W and Yun, Sangdoo and Song, Hyun Oh},
  journal={Advances in Neural Information Processing Systems},
  volume={38},
  pages={167563--167591},
  year={2026}
}

\clearpage
\appendix
% No layout overrides here on purpose: AAAI forbids changing type size, leading
% and spacing, so float placement, caption skip and all vertical spacing are
% left exactly as aaai2027.sty and the class set them.

\section*{LLM Usage Statement}
\label{app:llm_usage}
We used an LLM for writing assistance, including drafting and revising prose, and for routine coding support such as plotting and job scripts.
The authors designed and ran every experiment, checked every reported number against our own logs, and take full responsibility for the content of this paper.

\section*{How to Read This Appendix}
This appendix is meant to be read next to the main paper, so we recap only the notation.
Throughout, $\rho$ is the retain ratio, the fraction of dense \KV{} channels kept, so $\rho{=}0.7$ means $30\%$ compression. \emph{Iso-\KV{}} means every method is given the same $\rho$, and \emph{post-KD} means after the shared LoRA--KD recovery pass. $D$ is the head dimension and $\hat D{=}HD$ the model dimension over $H$ query heads, $H_{kv}$ counts \KV{} heads, $m_\ell$ is the per-head pair budget at layer $\ell$, and $\theta_{\text{base}}$ is the RoPE base frequency. Pair saliency turns up in two forms, the weight-scaled Fisher saliency $s_p$ that the analysis motivates and the root-Fisher aggregate $\sigma_p$ that the implementation actually computes.
The order runs from the method's limits to the measurements. \Cref{app:limitations} says where RAP loses. \Cref{app:rope_adoption} checks that the pair constraint binds in models people deploy. \Cref{app:method,app:cost_model} give the derivations. \Cref{app:repro} records how our numbers should be read, and it is the one section to consult before quoting any single cell of ours. \Cref{app:baselines,app:extended,app:efficiency} then take the baseline objections, the quality runs beyond the default model, and the efficiency measurements.
Tables numbered below~4 and figures numbered below~8 belong to the main paper. We cite them by number and do not reproduce them here.

\section{Limitations and Discussion}
\label[appendix]{app:limitations}
\label{sec:limitations}
We start with the places RAP loses. The budget we recommend is a trade-off, and this section gives its edges.

\noindent\textbf{Stronger compression favors low-rank.}
At $\rho{=}0.7$ RAP stays within about $5$ LongBench points of PaLU while using lower attention cost and needing no reconstruction (\Cref{tab:main_results}).
At $\rho{=}0.5$ that ordering breaks. PaLU degrades far more gracefully there, scoring roughly $48$ to $50$ on LongBench against RAP's $29$ to $33$.
The reason is structural rather than incidental. Pruning removes dimensions outright, so at half the budget the surviving pairs have to carry all of the positional information by themselves, while a low-rank factor is still free to fold information from the removed directions into the ones it keeps.
So we put $\rho{=}0.7$ forward as the practical trade-off, where the quality gap is small and the cost and reconstruction advantages are real, and we claim no dominance at aggressive budgets.

\noindent\textbf{The pair constraint is specific to RoPE.}
RAP's correctness argument rests on RoPE applying independent $2{\times}2$ rotations, which is what makes the binary expansion commute (\Cref{eq:rope-commute-motivation}).
Models with learned absolute embeddings, ALiBi, or other non-RoPE schemes do not have this structure, so the pair constraint is either meaningless or unnecessary there and RAP reduces to ordinary channel pruning.
Our checks cover Mistral and Qwen, scales from 3B to 14B, and a low-base RoPE control (\Cref{app:extended}). Every one of those is a RoPE model, so other positional-encoding families remain untested.

\noindent\textbf{Serving gains are context-dependent.}
End-to-end decode gains appear at 16k to 32k context at batch~1 (\Cref{app:efficiency}), because that is where the cache dominates the step.
At short context the projection GEMMs dominate instead, and an unfused reduced-dimension projection is not automatically faster than a dense one. On the memory-optimized path of \Cref{app:efficiency} RAP runs at $20.5$ tokens/s against $34.5$ for dense at batch~1.
Turning the saving into throughput inside a batched serving stack needs a fused attention kernel, and we have not written one. The Triton kernel we do provide fuses the RoPE gather rather than attention, which is a different thing. We report the memory capacity the smaller cache already buys. Fused attention is future work.

\noindent\textbf{Recovery is required at the operating point.}
Without any fine-tuning at all, RAP is usable only up to about $10\%$ compression. Pruning alone gives $8.85$ PPL at $\rho{=}0.9$ against $8.28$ for dense, then $11.65$ at $\rho{=}0.8$ and $28.9$ at $\rho{=}0.7$. Those three come from the pooled-budget variant, so they bound the recovery-free quality from the pessimistic side.
Getting to $\rho{=}0.7$ takes the LoRA--KD pass, which turns that $28.9$ into $9.12$ for roughly $9$ GPU-hours of work. That is cheap next to pretraining, but it is not free and it needs calibration data.
That is the trade we chose, and it is why the paper reports a recovered operating point instead of a recovery-free one.

\section{RoPE in Recent Open LLMs}
\label[appendix]{app:rope_adoption}
\Cref{sec:intro} argues that RoPE-safe pruning matters in practice. That argument only has force if the models people actually deploy rotate paired dimensions, so \Cref{tab:hybrid_rope} goes through the open models released in the year before submission and locates where the pair constraint binds in each.
Every one of them applies RoPE in its softmax-attention layers. Three do it in a modified form worth naming, because the modification changes where RAP applies rather than whether it applies. Llama-4 interleaves RoPE layers with global layers that carry no positional embedding at all (iRoPE), while Qwen3-Next and GLM-4.7 rotate only a fraction of each head's dimensions (partial rotary).
In both forms the rotated dimensions still arrive in $2{\times}2$ blocks. RAP therefore applies unchanged to the rotated part and falls back to ordinary channel pruning on the rest, which is the hybrid rule of \Cref{sec:rap-serving}. Working it through on the concrete cases, Llama-4-Scout places a NoPE layer every fourth layer, so the pair constraint binds on $36$ of its $48$ layers and the other $12$ are pruned channel-wise. GLM-4.7 rotates half of each head's dimensions, and Qwen3-Next rotates $64$ of $256$.
None of this disturbs the accounting. Each layer still keeps the same fraction $\rho$ of its \KV{} channels, so the cache, parameter, and FLOP scaling of \Cref{tab:rap_complexity} carries over untouched. The hybrid rule changes which dimensions may be dropped independently, not how many.
It would be easy to over-read this section. \Cref{tab:hybrid_rope} reads off released configurations. We ran nothing on these models. The weights are public, but pruning and recovering six more models was beyond the compute we had, so we claim no measured quality for any of them.

\begin{table}[t]
\centering
\small
\setlength{\tabcolsep}{3pt}
\begin{tabular*}{\linewidth}{@{\extracolsep{\fill}}l c c c@{}}
\toprule
\textbf{Model} & \textbf{Release} & \textbf{RoPE layers} & \textbf{Rot.\ dims/head} \\
\midrule
GLM-4.7 & Dec 2025 & all & 64/128 \\
Qwen3-Next & Sep 2025 & attn.\ only & 64/256 \\
GPT-OSS & Aug 2025 & all & all \\
Mistral-Small-3.2 & Jun 2025 & all & all \\
Llama-4-Scout & Apr 2025 & 36/48 & 128/128 \\
Gemma-3 & Mar 2025 & all & all \\
\bottomrule
\end{tabular*}
\caption{Where the RoPE-pair constraint binds in recent open LLMs, from their released configurations~\citep{glm47, qwen3next, gptoss, mistral32, llama4, farabet2025introducing}. Every model applies RoPE in its softmax-attention layers. Llama-4 interleaves NoPE layers (iRoPE), and GLM-4.7 and Qwen3-Next rotate only part of each head (partial rotary). RAP prunes the rotated dimensions in pairs and the rest channel-wise.}
\label{tab:hybrid_rope}
\end{table}

\section{Extended Analysis of RAP Design}
\label[appendix]{app:method}

\noindent\textbf{Second-order loss estimate.}
This section expands \Cref{prop:loss_estimate}, first connecting reduced RAP attention to a full-dimensional pruning perturbation and then deriving its pair-additive loss estimate.

\noindent\emph{Full-dimensional representation.}
Let $\E_k\in\mathbb{R}^{2m\times D}$ select the coordinates of the retained set $\mathcal{S}$ and define
\begin{equation}
\M_{\mathcal S}\triangleq\E_k^\top\E_k.
\end{equation}
The rows of $\E_k$ are distinct coordinate vectors, so $\E_k\E_k^\top=\mathbf{I}_{2m}$, while $\M_{\mathcal S}$ is a diagonal mask with ones on retained coordinates.
We also have
\begin{equation}
\widetilde\W_k\E_k
=\W_k\E_k^\top\E_k
=\W_k\M_{\mathcal S}.
\label{eq:masked-expansion}
\end{equation}
Thus expanding the reduced projection reconstructs the masked projection, not the original dense projection.

For completeness, let $\R_t$ be the full-dimensional RoPE rotation at position $t$, and let $\R_{t,\mathcal S}$ contain the rotation blocks retained by $\mathcal{S}$.
Because $\E_k$ selects complete RoPE pairs,
\begin{equation}
\E_k\R_t=\R_{t,\mathcal S}\E_k,
\qquad
\R_t\M_{\mathcal S}=\M_{\mathcal S}\R_t.
\label{eq:selection-rope-intertwine}
\end{equation}
The first equality is the precise intertwining relation that the main text calls ``commuting with RoPE''. The second holds because $\M_{\mathcal S}$ is constant on each $2{\times}2$ rotation block, so it commutes with the block-diagonal $\R_t$.
For full-dimensional query and key feature vectors $\mathbf{q},\mathbf{k}$ at positions $t,u$, it follows that
\begin{equation}
(\R_{t,\mathcal S}\E_k\mathbf{q})^\top
(\R_{u,\mathcal S}\E_k\mathbf{k})
=
(\R_t\mathbf{q})^\top\M_{\mathcal S}(\R_u\mathbf{k}).
\label{eq:masked-logit-equivalence}
\end{equation}
On the left is the reduced RAP dot product, the one implemented by $\widetilde\W_q=\W_q\E_k^\top$ and $\widetilde\W_k=\W_k\E_k^\top$. On the right is the full-dimensional dot product after zeroing the removed pairs. They are the same number.
Because RAP retains the original scale $1/\sqrt D$, the corresponding attention logits are identical.

\noindent\emph{Second-order estimate.}
The full-dimensional perturbation induced by pruning is
\begin{equation}
\Delta\W_k
\triangleq\W_k\M_{\mathcal S}-\W_k
=\W_k(\M_{\mathcal S}-\mathbf{I}_D).
\end{equation}
It is zero on retained pairs and equals $-\W_k^{(p)}$ on each removed pair $p\notin\mathcal S$.
A Taylor expansion of the calibration loss around the dense weights gives
\begin{equation}
\begin{split}
\Delta\mathcal L
={}&\langle\nabla\mathcal L,\Delta\W_k\rangle\\
&+\tfrac12\operatorname{vec}(\Delta\W_k)^\top
\nabla_{\W_k}^{2}\mathcal L\,
\operatorname{vec}(\Delta\W_k)
+R_3.
\end{split}
\label{eq:taylor-pruning}
\end{equation}
Assumption (i) of \Cref{prop:loss_estimate} neglects the linear term and higher-order remainder $R_3$ for this perturbation.
Assumption (ii) replaces the Hessian by the empirical diagonal Fisher of Eq.~\eqref{eq:fisher}, a standard saliency approximation~\citep{lecun-obd, molchanov2019importance, kwon2022fast}.
The diagonal approximation removes all cross-parameter, and hence cross-pair, curvature terms:
\begin{align}
\Delta\mathcal L
&\approx\tfrac12\sum_{n,i}F_{n,i}(\Delta\W_k)_{n,i}^{2}\nonumber\\
&=\tfrac12\sum_{p\notin\mathcal S}
\sum_{i\in p}\sum_nF_{n,i}(\W_k)_{n,i}^{2}
=\tfrac12\sum_{p\notin\mathcal S}s_p.
\label{eq:pair-additive-derivation}
\end{align}
Pair-preserving pruning is what defines the groups in this sum, and the additivity comes from the diagonal-curvature approximation.

\noindent\emph{Fixed-budget selection.}
For $|\mathcal S|=m$,
\begin{equation}
\sum_{p\notin\mathcal S}s_p
=\sum_{p\in\mathcal P}s_p-\sum_{p\in\mathcal S}s_p.
\end{equation}
The first term is constant, so minimizing the estimated removed saliency is equivalent to maximizing retained saliency, i.e., retaining the $m$ largest pair scores.
The derivation therefore provides a pair-additive objective for RAP's grouped selection.
RAP instantiates this objective with the practical root-Fisher aggregate $\sigma_p$ of Eq.~\eqref{eq:score}, and we measure what that substitution costs in the next paragraph. Layer allocation and the balanced $\K/\V$ split stay separate design choices.

\noindent\textbf{Root-Fisher versus weight-scaled saliency.}
\Cref{prop:loss_estimate} motivates the weight-scaled Fisher saliency $s_p$, while RAP ranks pairs by the cheaper root-Fisher aggregate $\sigma_p$. That substitution has to be accounted for, so we measured how far apart the two orderings actually are.
We measured it directly on all $256$ \KV{} heads of Llama-3-8B, computing both scores from the same cached Fisher and the dense $\W_k$ and comparing which pairs each one retains.
The two are not interchangeable: the mean Spearman correlation between the orderings is $0.67$, and the selected sets overlap by $85\%$ on average at $\rho{=}0.7$ ($75\%$ at $\rho{=}0.5$, $95\%$ at $\rho{=}0.9$), with fewer than $4\%$ of heads selecting an identical set at any ratio.
We did not re-run recovery under $s_p$ itself, so what we can say is narrower: swapping the root-Fisher aggregate for the much cruder weight magnitude moves recovered perplexity by $0.17$, inside the single-run tolerance (\Cref{tab:ablation_design}).
We therefore read the estimate as motivating \emph{pair-additive scoring at RoPE-pair granularity} rather than endorsing one saliency functional: the pair constraint carries the result, and the choice of score within that constraint is second order.

\begin{figure}[t]
\centering
\includegraphics[width=\linewidth]{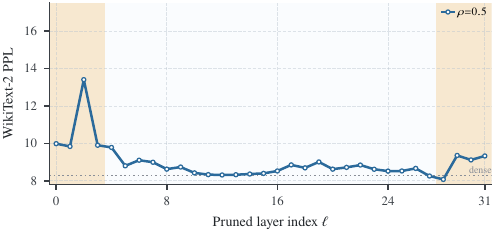}
\caption{Single-layer prune probe (Llama-3-8B): prune one layer, keep the rest dense. Shaded bands mark the first and last four layers.}
\label{fig:layer_probe}
\end{figure}

\noindent\textbf{Single-layer prune probe.}
\Cref{sec:allocation} gives layers different budgets on the grounds that they are not equally sensitive to pruning.
\Cref{fig:layer_probe} is the measurement behind that rule.
We prune exactly one layer to a target ratio, leave every other layer dense, and record WikiText-2 perplexity, which isolates the damage attributable to that layer.
The curve is strongly non-uniform.
At $\rho{=}0.5$ the first four layers average $10.8$ PPL and the last four average $9.0$, against $8.6$ for the middle of the network and $8.28$ for the dense model, and layer~2 alone reaches $13.4$.
A uniform budget therefore spends the same number of channels on layers that cost almost nothing to prune and on layers that dominate the total damage, which is what the layer-adaptive rule corrects.

\noindent\textbf{The ablated joint $\K$/$\V$ allocator.}
The allocator RAP does \emph{not} use keeps the layer-adaptive rule but lets $\K$ and $\V$ groups compete inside one Fisher pool, which is the natural choice if Fisher scores are read as comparable budgets.
They are not comparable in practice.
Calibration sequences are short, so $\V$ accumulates a larger Fisher mass than $\K$, and the pooled allocator responds by starving the key side: at $\rho{=}0.5$ it leaves $\K$ with about $21\%$ of the head dimension, against the $50\%$ a balanced split gives.
What that costs is long-context accuracy.
Under the same LoRA--KD recipe at $\rho{=}0.7$, the pooled allocator reaches $43.4$ on LongBench against $51.4$ for the balanced split (\Cref{tab:kvbalance}).
One caveat on that gap. An earlier partial evaluation of the same pooled checkpoint scored TriviaQA at $94.0$, above the dense model's own $90.0$. We treat that as anomalous and exclude it. Taking it instead would narrow the gap to about $3$ points. Under the reported reading the balanced split is ahead on all five tasks, and under the $94.0$ reading it is ahead on four.
RAP therefore keeps layer-adaptive budgets but fixes the $\K$/$\V$ ratio (\Cref{sec:ablation,tab:kvsplit}).

\begin{table}[t]
\centering
\small
\begin{tabular*}{\linewidth}{@{\extracolsep{\fill}}l cc@{}}
\toprule
\textbf{LongBench task} & \textbf{Pooled Fisher} & \textbf{Balanced $\K$/$\V$} \\
\midrule
TriviaQA & 70.1 & 82.8 \\
Qasper   & 21.8 & 35.9 \\
TREC     & 55.0 & 62.0 \\
SAMSum   & 39.7 & 40.6 \\
LCC      & 30.4 & 35.9 \\
\midrule
Average  & 43.4 & \textbf{51.4} \\
\bottomrule
\end{tabular*}
\caption{Why the $\K$/$\V$ budget is fixed rather than pooled (Llama-3-8B, $\rho{=}0.7$, post-KD, one run per column). Both columns share the layer-adaptive rule and the same LoRA--KD recipe and differ only in whether $\K$ and $\V$ compete in one Fisher pool. Per-task margins run from $0.9$ to $14.1$ points on $50$ examples each, so the average is the load-bearing number.}
\label{tab:kvbalance}
\end{table}

\section{Cost Model Derivation}
\label[appendix]{app:cost_model}
\Cref{tab:rap_complexity} states how cache, parameters, and FLOPs scale with $\rho$ for each family, and the claim that only RAP scales linearly in all three is central to \Cref{sec:efficiency}.
This section derives those entries.
We count one \KV{} head with input dimension $\hat D{=}HD$, so each projection is $\hat D\times D$, and we count only $\W_k,\W_v$ and the runtime work needed to produce the cached states.
For the low-rank rows this includes reconstruction, because that work is what iso-\KV{} accounting hides: SVD-LLM reconstructs both $\K$ and $\V$, while PaLU absorbs the $\V$-side factor into $\W_o$ and reconstructs only $\K$.
All entries are written as a multiple of the dense baseline, so the per-head convention does not affect the ratios.

\noindent\textbf{Baseline.}
Caching full $\K,\V$ costs $2SD$ per \KV{} head ($2SH_{kv}D$ over $H_{kv}$ heads under GQA), the two projections hold $2\hat DD{=}2HD^2$ parameters, and projecting $S$ tokens costs $4S\hat DD{=}4SHD^2$ FLOPs.

\noindent\textbf{SVD-LLM.}
Each of $\W_k,\W_v$ is factorized as $\A\B$ with rank $\rho D$, where $\A\in\mathbb{R}^{\hat D\times\rho D}$ and $\B\in\mathbb{R}^{\rho D\times D}$.
The cache holds the latents $\X\A_k,\X\A_v$, so it scales by $\rho$.
Parameters and FLOPs do not, because both factors stay online:
\begin{align}
\mathrm{Params} &= 2\bigl(\hat D\rho D+\rho D\cdot D\bigr)=\Bigl(\rho+\tfrac{\rho}{H}\Bigr)2HD^2,\\
\mathrm{FLOPs}  &= \underbrace{4S\hat D\rho D}_{\text{project}}+\underbrace{4S\rho D\cdot D}_{\text{reconstruct}}=\Bigl(\rho+\tfrac{\rho}{H}\Bigr)4SHD^2 .
\end{align}

\noindent\textbf{PaLU.}
Whitening plus absorption removes the $\V$-side reconstruction, so $\V$ keeps only $\A_v$ while $\K$ keeps both $\A_k$ and $\B_k$:
\begin{align}
\mathrm{Params} &= \underbrace{\hat D\rho D+\rho D\cdot D}_{\K}+\underbrace{\hat D\rho D}_{\V}=\Bigl(\rho+\tfrac{\rho}{2H}\Bigr)2HD^2,\\
\mathrm{FLOPs}  &= \underbrace{2S\hat D\rho D+2S\rho D\cdot D}_{\K}+\underbrace{2S\hat D\rho D}_{\V} \nonumber\\
                &= \Bigl(\rho+\tfrac{\rho}{2H}\Bigr)4SHD^2 .
\end{align}
The leftover $\B_k$ costs $\rho/2H$, which is $1.1\%$ of the dense $\K$ and $\V$ projections at $\rho{=}0.7$ and $H{=}32$, and it is what forces per-step reconstruction during decode. It is not what keeps PaLU well above the $\rho$ line in \Cref{fig:accuracy_fig}(D). That is $\W_q$, and the Scope paragraph below accounts for it.

\noindent\textbf{RAP.}
The retained columns form $\widetilde\W_k,\widetilde\W_v\in\mathbb{R}^{\hat D\times\rho D}$ and the expansions $\E_k,\E_v$ are binary index maps that fuse offline into $\W_q$ and $\W_o$ and therefore contribute no online parameters and no online FLOPs:
\begin{equation}
\begin{gathered}
\mathrm{Cache}=\rho\,2SD,\quad
\mathrm{Params}=\rho\,2HD^2,\\
\mathrm{FLOPs}=\rho\,4SHD^2 .
\end{gathered}
\end{equation}
All three scale as $\rho$, which is the linear row of \Cref{tab:rap_complexity}.

\noindent\textbf{Scope.}
This model isolates the \KV{}-projection cost of one head.
\Cref{tab:effsweep} counts the same quantities on the deployed models across the whole attention block, which is wider than the two projections above, so its absolute values differ. The shape is what carries over.
RAP's counted ratio equals $\rho$ to within $0.7$ percentage points at every budget, the residual being the rounding of a per-layer budget to whole pairs.
The two low-rank rows are instead affine in $\rho$, with an intercept no budget removes: fitting the five cells gives $0.40+0.60\rho$ for PaLU and $0.80+0.24\rho$ for SVD-LLM.
Those intercepts are not fitting artifacts. On Llama-3-8B the four attention projections are $\W_q$ and $\W_o$ at $16.78$M each and $\W_k,\W_v$ at $4.19$M each under GQA, so $\W_q$ alone is $0.400$ of the block and $\W_q$ with $\W_o$ is $0.800$, which are exactly the two intercepts.
The reason is the fusion each method can perform. PaLU absorbs its $\V$-side factor into $\W_o$, but RoPE blocks the $\K$-side factor from entering $\W_q$, so $\W_q$ stays at full width. SVD-LLM absorbs neither, so $\W_q$ and $\W_o$ both stay. RAP fuses into both and leaves nothing behind, and that is why its row is proportional to $\rho$ instead of affine in it.
So the cost a low-rank method cannot compress is not the small leftover factor but the query projection it was never able to shrink.
Wall-clock is a further step removed: GQA sharing and kernel details move the constants, which is why the latency ratios later in this appendix do not follow the FLOP ratios exactly.

\section{Reproducibility}
\label[appendix]{app:repro}
Setup hyper-parameters are in \Cref{sec:setup}.
Here we record what affects how the numbers should be read.
Unless labeled otherwise, every figure and table comes from our own pipeline under the shared Fisher allocation and LoRA--KD recipe. At a given model and budget the recipe is identical across methods, which is what keeps a method comparison from turning into a comparison of recovery effort. Across models it is not identical, and we list the deviations for the 3B, 14B, and bf16 runs below.
The code for pruning, recovery and evaluation accompanies this submission as a source-only supplementary archive, with the entry point for each kind of measurement named in its README. It will be released under a permissive open-source license on publication.
All models (Llama, Mistral, Qwen) are used under their released licenses, and all evaluation data is public: WikiText-2~\citep{wikitext2}, the six zero-shot tasks (OpenBookQA~\citep{openbookqa}, HellaSwag~\citep{hellaswag}, PIQA~\citep{piqa}, ARC-Easy and ARC-Challenge~\citep{arc}, WinoGrande~\citep{winogrande}), and LongBench~\citep{longbench}.

\noindent\textbf{How precise RAP's own number is.}
Each RAP cell of \Cref{tab:main_results} comes from two independent recovery runs that differ only in the training seed, each of them evaluated two or three times. The printed value is the mean over those evaluations.
Between seeds the recovery is very reproducible: the two seed means differ by $0.006$ PPL at $\rho{=}0.7$ on Llama and $0.001$ on Mistral. Between evaluations of one seed the spread is larger, up to $0.30$ PPL on Llama and $0.03$ on Mistral, and it is that larger figure which sets the resolution of a Llama comparison here at roughly a third of a point.
Cells elsewhere in this appendix are single runs. Earlier single runs sit up to $0.55$ PPL above the later repeated means on Llama and are indistinguishable from them on Mistral, so a single-run cell should be read with a ${\sim}0.5$-PPL tolerance.
One consequence is visible in \Cref{tab:pplsweep,tab:pplsweep_mistral}: at $\rho\ge0.8$ the recovered PaLU cells sit at or just below the dense baseline. When a compressed model recovers past its teacher on the very corpus recovery used, what that shows is mild domain fitting. It does not show compression helping. That is why no conclusion here rests on a difference below the tolerance.

\noindent\textbf{How precise KVPruner's number needs to be.}
The collapsed baseline is far noisier, $78.9\pm12.8$ on Mistral at $\rho{=}0.7$, and it is also non-monotonic in the budget on both models: $935, 152, 397, 186, 94.2$ on Llama and $26.7, 36.0, 78.9, 33.3, 27.1$ on Mistral.
That is what breaking pairs does. Once most surviving channels have lost their partner, the post-KD result is set by how the collapse happens, and the number of channels that survive stops mattering, so it need not improve as the budget grows.
On Mistral its worst point lands exactly on $\rho{=}0.7$, the budget we recommend, and on Llama the sweep is erratic enough that no single budget is safe to quote. So the comparison worth making is against its best point anywhere in the sweep: $94.2$ on Llama and $26.7$ on Mistral.
RAP is still $10\times$ better than that on Llama and $4\times$ on Mistral. We make the point without a significance test, because with two seeds per method and collapsed, heavily skewed baseline values a $t$-test would be the weakest link in the argument: taking KVPruner at its best logged evaluation and RAP at its worst preserves the same ordering and the same order of magnitude on Llama, which needs no distributional assumption.
No conclusion here needs the KVPruner mean to be precise.

\noindent\textbf{Where the baseline's noise comes from.}
We first read that noise as an evaluation artifact. It is not one. Our two logging paths agree to four significant figures whenever they load the same adapter (at Mistral $\rho\in\{0.6,0.8,0.9\}$ they differ by less than $0.005$ PPL), so evaluation is not the variable.
The variable is recovery. Two seed-isolated LoRA--KD runs of KVPruner at Mistral $\rho{=}0.7$ give $64.8$ and, across two evaluations of the second seed, $82.2$ and $89.8$ PPL. A third run outside the seed-isolated set reaches $127.0$. So runs differing only in seed span a factor of two, and the $78.9$ we report is the mean of the three seed-isolated evaluations, with the $127.0$ excluded.
We report the mean of the seed-isolated runs ($78.9$). On the same model RAP's two seeds land $0.001$ PPL apart, and even its spread across evaluations is only $0.03$. That asymmetry, a factor of two against a thousandth of a point, is what the comparison actually rests on.

\noindent\textbf{Calibration and domain sensitivity.}
Fisher scores come from 32 sequences of 1024 tokens of WikiText-2 train and the same corpus is used for LoRA--KD, so in-domain evaluation could in principle flatter the recovery.
A C4-en~\citep{c4} KD control on Llama at $\rho{=}0.7$ reaches $9.67$ PPL against $9.12\pm0.06$ in-domain, a gap of about half a point, so recovery is not an artifact of overlap between the KD corpus and the evaluation corpus.
We did not sweep the calibration size or the KD temperature, so we make no claim about them beyond the defaults reported here.

\noindent\textbf{Recovery schedule.}
LoRA--KD trains with AdamW at learning rate $10^{-4}$ under a reduce-on-plateau schedule, with global batch size 128 via gradient accumulation, sequence length 256, gradient-norm clipping, and up to 20 epochs on WikiText-2 train, keeping the checkpoint with the best validation perplexity.
The 3B and 14B students are less stable under this default and use learning rate $2{\times}10^{-5}$ with clip 0.3 (8 epochs for 14B).
One recovery run costs roughly $9$ GPU-hours on a single A100, most of which went to our own periodic full-corpus perplexity evaluation for checkpoint selection and only a small part to the LoRA update itself.

\noindent\textbf{Precision and hardware.}
All accuracy numbers are evaluated in fp32 to avoid the dtype mismatch that pruned models can hit when RoPE tables and weights disagree.
Latency uses fp16 with CUDA-event timing, warmup and repeats, on one card at a time, following the attention-module protocol of \Cref{fig:efficiency}.
At Mistral $\rho\in\{0.6,0.8,0.9\}$ the RAP student used the same schedule in bf16, while the PaLU and KVPruner cells of those rows are fp32. An fp32 canary at $\rho{=}0.6$ matched the ordering, and the resulting points lie on the same monotone curve as the fp32 points in \Cref{tab:pplsweep_mistral}.
The Qwen2.5-14B student also trains in bf16, since an fp32 student of that size does not fit alongside its teacher on one 80GB GPU.

\section{Extended Baseline Comparison}
\label[appendix]{app:baselines}

\noindent\textbf{KVPruner and why its failure is not a setup artifact.}
KVPruner numbers come from a same-framework reimplementation that shares RAP's Fisher scores, iso-\KV{} budget, and KD recipe, and differs only in selecting individual channels instead of whole RoPE pairs.
That single-variable design is what lets us attribute the gap to pruning granularity, but it invites two objections, and we address both.
The first is that we might be handicapping KVPruner with a naive RoPE policy.
\Cref{tab:steelman} answers it with an index-aware steelman that rotates each surviving half-pair with its original frequency instead of a re-indexed one.
The steelman helps at $\rho{=}0.7$, moving PPL from $397$ to $94$, and still trails RAP by an order of magnitude after the same KD, which says the failure mode is the orphaned half-pairs themselves rather than index bookkeeping.
At $\rho{=}0.5$ it does not help at all ($952$ against $935$): by then so few pairs are intact that rotating the survivors at the right frequency has nothing left to preserve.
The second objection is that the published method worked, so our reproduction must be wrong.
It behaves differently in the regime it was published in, and that difference is the point.
On Llama-2-7B, which is MHA with base $10^4$, KVPruner does respond to recovery: $12833$ pre-recovery becomes $60.4$ post-KD at $\rho{=}0.5$, a factor of over $200$. That is still far from usable, and we do not claim to have reproduced a deployable KVPruner. We do reproduce the qualitative fact that recovery moves it on MHA. On the same model RAP goes from $137.9$ to $6.50$.
The collapse appears on the modern high-base GQA models (Llama-3, Mistral-7B-v0.3, Qwen2.5), which is exactly the regime this paper is about, and the RoPE-base control in \Cref{fig:rope_base_fig} isolates the base frequency from the MHA/GQA confound.

\begin{table}[t]
\centering
\small
\begin{tabular*}{\linewidth}{@{\extracolsep{\fill}}l cc@{}}
\toprule
\textbf{Variant} & \textbf{retain 0.5} & \textbf{retain 0.7} \\
\midrule
KVPruner (vanilla) & 935 & 397 \\
KVPruner (index-aware) & 952 & 94 \\
\textbf{RAP (balanced)} & \textbf{10.82} & \textbf{9.12} \\
\bottomrule
\end{tabular*}
\caption{Index-aware KVPruner steelman (Llama-3-8B, post-KD PPL$\downarrow$). The RAP row gives the repeated-run means of \Cref{tab:pplsweep} and the KVPruner rows are single runs. The ratios are $10\text{--}88\times$ on these values. The main paper quotes $10\text{--}84\times$, computed against the single-run RAP figures.}
\label{tab:steelman}
\end{table}

\noindent\textbf{PaLU.}
All PaLU rows use the official decomposition together with the same LoRA--KD recipe as RAP and KVPruner, which gives Llama $8.59$ at $\rho{=}0.7$.
We first tried an in-house reimplementation and it recovered poorly under this KD. Rather than report the weaker number, we adopted the official pipeline, so PaLU appears in our tables at its strongest.
On latency, PaLU's paper reports a large speedup that comes primarily from a custom kernel fusing the projections, reconstruction, RoPE, and attention.
Our comparison in \Cref{fig:efficiency} is at the attention-module level with the same measurement protocol for both methods, so what it measures is the reconstruction work RAP removes, and kernel-engineering differences drop out of the comparison.
A fused implementation would lift both methods. It would not remove PaLU's per-step $\K$ reconstruction, which is the structural term.

\noindent\textbf{EliteKV and MatryoshkaKV.}
Both are omitted from \Cref{tab:main_results} because they do not converge under the shared LoRA--KD recipe (\Cref{tab:nonconverging}).
We did not retune recovery for these two, yet we did switch PaLU to its authors' pipeline after our own reimplementation recovered poorly. That concession went to the strongest baseline, which makes our claims harder to support rather than easier. For EliteKV and MatryoshkaKV it means \Cref{tab:nonconverging} is a lower bound. Our recipe does not recover them, which is not the same as saying nothing would.
EliteKV shares one low-rank latent between $\K$ and $\V$, and GQA gives that shared latent very little to work with: pre-recovery PPL is $6459$ on Llama-3 at $\rho{=}0.7$ against $456$ on MHA Llama-2, and it then diverges under our KD.
Its native recipe is full uptraining rather than a LoRA pass, so we report the pre-recovery number and flag the mismatch instead of claiming it fails outright.
MatryoshkaKV is activation-side, so it leaves $\W_k,\W_v$ untouched and does not reduce parameters or projection FLOPs (\Cref{tab:capability}). Under the shared KD it degrades on three of the four points we ran.

\begin{table}[t]
\centering
\small
\setlength{\tabcolsep}{3pt}
\begin{tabular*}{\linewidth}{@{\extracolsep{\fill}}l l c cc@{}}
\toprule
\textbf{Method} & \textbf{Model} & \textbf{retain} & \textbf{Pre-rec.} & \textbf{Post-KD} \\
\midrule
EliteKV & Llama-3-8B & 0.5 & 55281 & div. \\
EliteKV & Llama-3-8B & 0.7 & 6459 & div. \\
MatryoshkaKV & Llama-3-8B & 0.5 & 784 & 491 \\
MatryoshkaKV & Llama-3-8B & 0.7 & 213 & 624 \\
MatryoshkaKV & Mistral-7B & 0.5 & 8.92 & 19.1 \\
MatryoshkaKV & Mistral-7B & 0.7 & 8.90 & 26.2 \\
\bottomrule
\end{tabular*}
\caption{Non-converging baselines under shared LoRA--KD (WikiText-2 PPL$\downarrow$). ``div.''${}={}$diverged.}
\label{tab:nonconverging}
\end{table}

\noindent\textbf{Low-rank projection methods we did not run.}
Two further low-rank \KV{} methods sit close to our setting. We did not evaluate either, but the RoPE argument of \Cref{sec:motivation} bears directly on both.
Eigen Attention (Saxena et al., Findings of EMNLP 2024) projects keys and values onto the leading eigenvectors of their covariance. It reaches the obstruction we describe and states it directly: LLMs with RoPE ``are trained with a fixed dimensional $R$, making them incompatible with any modification to the embedding dimension of the keys or queries''. Its remedy is to leave the query full-rank and transform the key back to high dimension before applying the rotation, which is exactly the per-step reconstruction RAP removes. An independent method hit the same obstruction and paid the same reconstruction cost, so it is not an artifact of how we set up PaLU.
OjaKV (Zhu et al., arXiv:2509.21623) adapts its projection basis online with Oja's rule while keeping the first and most recent tokens full-rank. Because the basis changes during generation, its expansion cannot be folded into $\W_q$ even in principle: offline fusion needs a fixed expansion, which is what makes RAP's binary, position-independent selection matrix the useful object here. What OjaKV buys instead is a basis fitted to the actual context, and that is orthogonal to pair alignment, so the two could compose on the tokens it does compress.
Both readings follow from what the methods do, not from anything we measured, and we make no quality claim about either.

\noindent\textbf{Whole-head pruning.}
Dropping entire \KV{} heads, together with the grouped $\Q$ heads and the matching $\W_o$ columns, is RoPE-safe because a head is a whole number of pairs.
Its problem is granularity: the budget moves in steps of $1/H_{kv}$, which is $12.5\%$ on Llama-3-8B, so it cannot express $\rho{=}0.7$ at all.
Where it can be compared, the pair constraint costs nothing: after the same KD, RAP reaches $10.82$ against head pruning $10.94$ at $\rho{=}0.5$ on Llama, and $6.39$ against $6.43$ on Mistral, both differences inside the tolerance of a single run.
Above that budget its nearest expressible option caches $0.75\times$, where it reaches $9.03$ on Llama and $5.63$ on Mistral. RAP brackets that point on both sides: at $\rho{=}0.8$ it reaches $8.59$ (\Cref{tab:pplsweep}) for $7\%$ more cache, and at $\rho{=}0.7$ it spends $7\%$ less cache at a perplexity difference of $0.09$, which is well inside the single-run tolerance.
Neither point dominates the other. What the coarse grid cannot do is land on the budget a deployment actually needs, and that is what the pair unit buys.

\section{Extended Quality Evaluation}
\label[appendix]{app:extended}
\Cref{sec:setup} claims that the findings are not specific to Llama-3-8B.
This section reports the runs behind that claim: the Mistral counterpart of the main sweep, the scale sweep, the RoPE-base control, the design ablations quoted in \Cref{sec:ablation}, a long-context retrieval probe, and the composition with 4-bit weights.

\begin{figure*}[!t]
\centering
\includegraphics{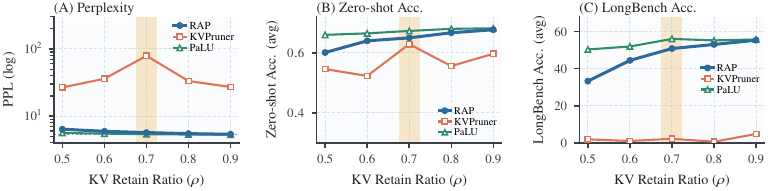}
\caption{Iso-\KV{} comparison of accuracy on Mistral-7B-v0.3 (post-KD), mirroring \Cref{fig:accuracy_fig}. The attention-cost panel is architecture-analytic and identical to \Cref{fig:accuracy_fig}(D).}
\label{fig:mistral_accuracy}
\end{figure*}

\begin{figure*}[!t]
\centering
\includegraphics[width=\textwidth]{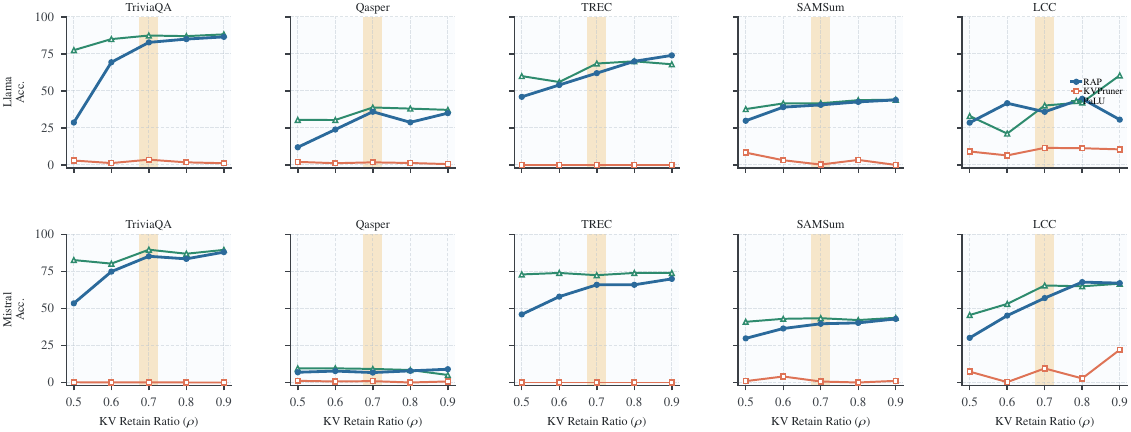}
\caption{LongBench task matrix (post-KD). Rows are Llama / Mistral. Columns are TriviaQA, Qasper, TREC, SAMSum, LCC.}
\label{fig:longbench_full}
\end{figure*}

\noindent\textbf{Mistral.}
\Cref{fig:mistral_accuracy} repeats the main-text sweep on Mistral-7B-v0.3 and reproduces the two findings the paper rests on: KVPruner collapses on LongBench at every budget, and RAP closes on PaLU as the budget relaxes, reaching $5.69$ against $5.38$ perplexity and $50.9$ against $56.1$ LongBench at $\rho{=}0.7$.
It also reproduces the limitation of \Cref{app:limitations} in sharper form: at $\rho{=}0.5$ PaLU leads LongBench by $17$ points ($50.4$ against $33.3$), so the aggressive-budget concession we make on Llama is not milder on Mistral.
Numeric perplexity for both models is in \Cref{tab:pplsweep,tab:pplsweep_mistral}, and \Cref{fig:longbench_full} breaks LongBench into its five tasks for both models, which shows the collapse is uniform across task types and not driven by one task.

\begin{table}[t]
\centering
\small
\setlength{\tabcolsep}{4pt}
\begin{tabular*}{\linewidth}{@{\extracolsep{\fill}}c ccc@{}}
\toprule
\textbf{retain} & \textbf{RAP} & \textbf{KVPruner} & \textbf{PaLU} \\
\midrule
Dense & \multicolumn{3}{c}{8.28} \\
\midrule
0.5 & 10.82 & 935 & 8.87 \\
0.6 & 9.69 & 152 & 8.64 \\
0.7 & 9.12 & 397 & 8.59 \\
0.8 & 8.59 & 186 & 8.20 \\
0.9 & 8.32 & 94.2 & 8.20 \\
\bottomrule
\end{tabular*}
\caption{Post-KD PPL (Llama-3-8B). $\rho{=}0.5$ and $0.7$ are means over two recovery seeds, $\rho{=}0.9$ is the mean of two evaluations of one seed, and $\rho{=}0.6$ and $0.8$ are single runs, to be read with the ${\sim}0.5$-PPL tolerance of \Cref{app:repro}.}
\label{tab:pplsweep}
\end{table}
\begin{table}[t]
\centering
\small
\setlength{\tabcolsep}{4pt}
\begin{tabular*}{\linewidth}{@{\extracolsep{\fill}}c ccc@{}}
\toprule
\textbf{retain} & \textbf{RAP} & \textbf{KVPruner} & \textbf{PaLU} \\
\midrule
Dense & \multicolumn{3}{c}{5.32} \\
\midrule
0.5 & 6.39 & 26.7 & 5.61 \\
0.6 & 5.94 & 36.0 & 5.49 \\
0.7 & 5.69 & 78.9 & 5.38 \\
0.8 & 5.49 & 33.3 & 5.32 \\
0.9 & 5.37 & 27.1 & 5.30 \\
\bottomrule
\end{tabular*}
\caption{Post-KD PPL (Mistral-7B-v0.3). $\rho{=}0.5$ and $0.7$ are means over two recovery seeds. At $\rho\in\{0.6,0.8,0.9\}$ the RAP student was trained in bf16 for the reason given in \Cref{app:repro}, while the PaLU and KVPruner cells of those rows are fp32. The fp32 canary at $\rho{=}0.6$ is the control for that asymmetry.}
\label{tab:pplsweep_mistral}
\end{table}

\noindent\textbf{Scale.}
\Cref{tab:generality} sweeps model scale at the operating point.
The gap to KVPruner carries across scale on perplexity, where RAP is $12$ to $44\times$ better at all three sizes. On LongBench we have the comparison only at 8B, $51.4$ against $3.4$. So the cross-scale evidence that the pair constraint matters is a perplexity result, and we do not extend the LongBench margin to 3B or 14B.
Against PaLU the picture is the one \Cref{app:limitations} concedes: PaLU keeps a perplexity edge at both scales where we can run it, $0.53$ PPL at 8B and $0.76$ at 14B, and we do not claim to close it.
We report LongBench for 3B and, in full, for Mistral, but leave the 14B cell empty. The 14B evaluations hit a recurring fault: one logged perplexity for that checkpoint is bit-identical to its pre-recovery value, and one of two LongBench runs is uniformly lower across all five tasks ($35.2$ against $56.2$). Both signatures say the adapter was not applied in those runs. For perplexity the two cases are distinguishable, so we report the recovered value, $6.70$. For LongBench they are not, so we report no RAP value there. The PaLU and KVPruner cells at 14B are unaffected by that fault ($52.9$ and $5.2$), and we leave them out only because a baseline column with no method column to compare against would invite exactly the comparison we are declining to make. No claim in the paper rests on the 14B LongBench row.

\begin{table}[t]
\centering
\small
\setlength{\tabcolsep}{2.5pt}
\begin{tabular*}{\linewidth}{@{\extracolsep{\fill}}l ccc ccc@{}}
\toprule
& \multicolumn{3}{c}{\textbf{PPL}$\downarrow$} & \multicolumn{3}{c}{\textbf{LongBench}$\uparrow$} \\
\cmidrule(r){2-4}\cmidrule(l){5-7}
\textbf{Model} & \textbf{RAP} & KVP. & PaLU & \textbf{RAP} & KVP. & PaLU \\
\midrule
3B  & 9.53 & 160.3 & --- & 39.2 & --- & --- \\
8B  & 9.12 & 397 & 8.59 & 51.4 & 3.4 & 55.3 \\
14B & 6.70 & 82.1 & 5.94 & --- & --- & --- \\
\bottomrule
\end{tabular*}
\caption{Scale at $\rho{=}0.7$ (post-KD): Llama-3.2-3B, Llama-3-8B, Qwen2.5-14B. PPL is a two-seed mean at 8B and a single run at 3B and 14B, and LongBench is the 5-task average. ``---'' marks cells we do not report: PaLU has no 3B configuration, we did not run KVPruner on LongBench at 3B, and the 14B LongBench row is withdrawn (see text).}
\label{tab:generality}
\end{table}

\begin{figure}[t]
\centering
\includegraphics[width=\linewidth]{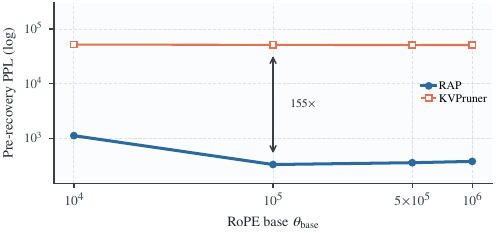}
\caption{RoPE-base sweep on fixed Llama-3-8B ($\theta_{\text{base}}{=}10^4$--$10^6$, $\rho{=}0.5$, pre-recovery).}
\label{fig:rope_base_fig}
\end{figure}

\noindent\textbf{RoPE base.}
The Llama-2 versus Llama-3 comparison in \Cref{app:baselines} moves two things at once, the base frequency and MHA versus GQA, so it cannot say which one matters.
\Cref{fig:rope_base_fig} separates them by overriding only $\theta_{\text{base}}$ on a fixed Llama-3-8B at $\rho{=}0.5$ and measuring pre-recovery perplexity.
The answer is that the base frequency is not the variable: KVPruner sits near $51000$ at every base from $10^4$ to $10^6$, so orphaning half-pairs is equally destructive in the regime the method was published in.
RAP moves somewhat with the base ($1120$ at $10^4$ against $330$--$380$ above it), so it sits between $45\times$ and $155\times$ below KVPruner depending on the base.
What differs between Llama-2 and Llama-3 is therefore attention layout. Rotation speed is not the variable: at the same base $10^4$, KVPruner's pre-recovery perplexity is $12833$ on MHA Llama-2 and near $51000$ on GQA Llama-3.
GQA is the aggravating factor, because fewer \KV{} heads mean each surviving pair is shared by more queries, and both numbers are pre-recovery, so this figure speaks to the damage and not to what recovery can repair.

\noindent\textbf{Design ablations.}
The numbers quoted in \Cref{sec:ablation} are collected in \Cref{tab:ablation_design}, and they should be read against the ${\sim}0.5$-PPL tolerance of single runs stated above.
Both of the choices people would expect to matter move perplexity by less than that resolution: scoring by $0.17$ (Fisher $9.35$, weight magnitude $9.52$) and the LoRA rank by at most $0.17$ ($9.52/9.35/9.37$ at $r{=}4/8/16$).
Neither choice is load-bearing. It is why RAP can score with the cheaper root-Fisher aggregate $\sigma_p$ rather than the weight-scaled Fisher saliency that the analysis of \Cref{sec:rap-theory} motivates, and why recovery does not need a tuned rank.
The $\K$/$\V$ split is the one design choice that leaves the noise floor: giving $\K$ $0.65$ of the channels costs $2.24$ PPL against the balanced split ($13.61$ against $11.37$ in \Cref{tab:kvsplit}), while the $0.35$ arm sits $0.27$ away and is not separable from it.
We take the balanced split because the $\K$-heavy side is clearly worse and because the balanced budget is what recovers long context (\Cref{tab:kvbalance}). Perplexity on its own would never separate $0.50$ from $0.35$.
What pruning achieves on its own and what recovery adds are separated in \Cref{fig:ablation_allocation}, whose prune-only bars share the pooled $\K$/$\V$ budget instead of RAP's balanced default, so the gap they show understates what recovery adds.

\begin{table}[t]
\centering
\small
\setlength{\tabcolsep}{4pt}
\begin{tabular*}{\linewidth}{@{\extracolsep{\fill}}l c c@{}}
\toprule
\textbf{Design choice} & \textbf{Setting} & \textbf{PPL$\downarrow$} \\
\midrule
\multirow{2}{*}{Pair scoring} & Fisher (default) & 9.35 \\
 & Weight magnitude & 9.52 \\
\midrule
\multirow{3}{*}{LoRA rank} & $r{=}4$ & 9.52 \\
 & $r{=}8$ (default) & 9.35 \\
 & $r{=}16$ & 9.37 \\
\bottomrule
\end{tabular*}
\caption{Design ablations at $\rho{=}0.7$ (Llama-3-8B, post-KD, single-seed). The $\K$/$\V$ split is in \Cref{tab:kvsplit}. Scoring and allocation are also plotted in \Cref{fig:ablation_allocation}, whose bars use the pooled $\K$/$\V$ budget and so do not match the balanced values here.}
\label{tab:ablation_design}
\end{table}
\begin{table}[t]
\centering
\small
\begin{tabular*}{\linewidth}{@{\extracolsep{\fill}}c c@{}}
\toprule
\textbf{$\K$ share} & \textbf{PPL$\downarrow$} \\
\midrule
0.35 & 11.64 \\
0.50 & 11.37 \\
0.65 & 13.61 \\
\bottomrule
\end{tabular*}
\caption{$\K$/$\V$ split at $\rho{=}0.5$ (Llama-3-8B, post-KD). All three arms come from one shared earlier recovery run, so the $0.50$ cell reads $11.37$ where \Cref{tab:pplsweep} reports $10.82$ for the same configuration under the repeated runs.}
\label{tab:kvsplit}
\end{table}

\noindent\textbf{Long-context retrieval.}
A needle-in-a-haystack probe turns out to be saturated at the depths an 8k-window model supports: at $\rho{=}0.7$ the dense model, RAP, and PaLU all retrieve every needle at 4k and 8k, and the only misses anywhere are RAP's at $\rho{=}0.5$, which retrieves $12$ of $15$ at 8k.
We omit 16k and 32k because the dense baseline scores zero there, so with no discriminating range left we rely on LongBench (\Cref{fig:longbench_full}) instead.

\noindent\textbf{Composing with 4-bit weights.}
\Cref{tab:quant4bit} gives the numbers behind \Cref{fig:quant_summary} on both models. The quantizer is round-to-nearest at 4 bits, per-tensor and symmetric, applied after recovery with the adapters merged. It is deliberately plain, and it costs the dense model a great deal on its own ($8.28$ to $14.75$), so what this table tests is whether pruning and quantization interact badly. How good 4-bit quantization can be on its own is a separate question we are not asking.
The two costs stack roughly additively down to $\rho{=}0.8$ and compound below it. At $\rho{=}0.9$ RAP is indistinguishable from the 4-bit dense baseline on both models ($14.54$ against $14.75$, $6.54$ against $6.41$), which given the size of the quantizer's own penalty says only that $10\%$ pruning adds nothing on top of it. At $\rho{=}0.7$ it falls $1.71$ and $1.87$ PPL behind. At $\rho{=}0.5$ the two costs compound and quality falls away.
On the same checkpoint the gap to dense widens from $0.54$ PPL in fp32 to $1.71$ at 4~bits on Llama, so composition is not free here.
The composition is a property of the recovered model rather than of pruning alone: the same checkpoints under the same quantizer reach $127.6$ PPL at $\rho{=}0.7$ without the LoRA--KD pass against $16.5$ with it.
We omit the low-rank baselines here because the quantized runs we have for them lack that shared pass, so the comparison would not be single-variable.

\begin{table}[t]
\centering
\small
\setlength{\tabcolsep}{2.5pt}
\begin{tabular*}{\linewidth}{@{\extracolsep{\fill}}l cc cc@{}}
\toprule
& \multicolumn{2}{c}{\textbf{Llama-3-8B}} & \multicolumn{2}{c}{\textbf{Mistral-7B}} \\
\cmidrule(r){2-3}\cmidrule(l){4-5}
\textbf{4-bit weights} & PPL$\downarrow$ & ZS$\uparrow$ & PPL$\downarrow$ & ZS$\uparrow$ \\
\midrule
Dense baseline & 14.75 & 0.602 & 6.41 & 0.649 \\
\midrule
RAP $\rho{=}0.9$ & 14.54 & 0.602 & 6.54 & 0.644 \\
RAP $\rho{=}0.8$ & 14.92 & 0.588 & 6.98 & 0.636 \\
RAP $\rho{=}0.7$ & 16.46 & 0.575 & 8.28 & 0.624 \\
RAP $\rho{=}0.6$ & 20.04 & 0.571 & 9.56 & 0.596 \\
RAP $\rho{=}0.5$ & 69.75 & 0.517 & 25.63 & 0.541 \\
\midrule
\multicolumn{5}{@{}l}{\textit{without LoRA--KD, same checkpoints and quantizer}} \\
RAP $\rho{=}0.7$ & 127.6 & 0.380 & 82.07 & 0.468 \\
\bottomrule
\end{tabular*}
\caption{RAP under 4-bit weight quantization (RTN, per-tensor, symmetric). ZS is the six-task average of \Cref{sec:setup}. The last row repeats $\rho{=}0.7$ without the recovery pass, which is what the composition depends on. These runs predate the balanced $\K$/$\V$ default and use the pooled-budget checkpoints, whose fp32 perplexity at $\rho{=}0.7$ is $8.82$. The comparisons in the text are against that value, not the balanced $9.12$.}
\label{tab:quant4bit}
\end{table}

\section{Extended Efficiency Evaluation}
\label[appendix]{app:efficiency}
This section gives the measurements behind \Cref{sec:efficiency}: the attention-cost sweep, the matched-compute comparison, end-to-end decode, the RoPE kernel microbenchmark, and serving memory.
The attention-module latencies of \Cref{fig:efficiency} isolate the component RAP changes, and they are timed with CUDA events, warmup, and repeats. But over 4k to 32k of context at batch~1 that component is only $0.8$ to $2.0$\,ms against a full decode step of $24$ to $70$\,ms. It is under $4\%$ of the step, so a ratio taken on it is sensitive to kernel and driver detail. The claims in this appendix therefore rest on the two quantities that are stable at this scale: counted parameters and FLOPs, which are exact, and end-to-end throughput, which is what a user sees.

\noindent\textbf{Attention cost and matched compute.}
\Cref{tab:effsweep} counts attention parameters and FLOPs as a fraction of dense across $\rho$, which is the deployed-model check on the cost model of \Cref{app:cost_model}.
RAP tracks the $\rho$ line to within $0.7$ percentage points at every budget while PaLU and SVD-LLM stay above it by a margin that grows as compression increases, because their cost is affine in $\rho$ rather than proportional to it (\Cref{app:cost_model}).
\Cref{tab:isoparam} uses that gap to ask a fairer question than iso-\KV{} alone: if the two methods are given the same attention compute, which one keeps more quality?
At matched ${\approx}0.70\times$ attention, RAP runs at $\rho{=}0.7$ while PaLU must drop to $\rho{=}0.5$.
PaLU keeps a perplexity lead of at most $0.25$ there, while RAP caches more ($0.70\times$ against $0.52\times$). On LongBench the ordering reverses, clearly on Llama ($51.4$ against $47.7$) and only nominally on Mistral ($50.9$ against $50.4$), where half a point on $50$ examples per task is inside the noise this appendix warns about elsewhere.

\begin{table}[t]
\centering
\small
\setlength{\tabcolsep}{3.5pt}
\begin{tabular*}{\linewidth}{@{\extracolsep{\fill}}c ccc@{}}
\toprule
\textbf{retain} & \textbf{RAP} & \textbf{PaLU} & \textbf{SVD} \\
\midrule
0.9 & \textbf{0.906} & 0.942 & 1.024 \\
0.8 & \textbf{0.797} & 0.881 & 0.999 \\
0.7 & \textbf{0.703} & 0.824 & 0.976 \\
0.6 & \textbf{0.594} & 0.763 & 0.951 \\
0.5 & \textbf{0.500} & 0.702 & 0.926 \\
\bottomrule
\end{tabular*}
\caption{Attention parameters and FLOPs as a fraction of dense, counted on the deployed Llama-3-8B models. Only RAP sits on the $\rho$ line at every budget.}
\label{tab:effsweep}
\end{table}

\begin{table}[t]
\centering
\small
\setlength{\tabcolsep}{3pt}
\begin{tabular*}{\linewidth}{@{\extracolsep{\fill}}l cc cc@{}}
\toprule
& \textbf{\KV{}} & \textbf{Par.} & \textbf{PPL} & \textbf{LB} \\
\midrule
\multicolumn{5}{@{}l}{\textit{Llama-3-8B}} \\
\textbf{RAP} $0.7$ & \textbf{0.70}$\times$ & 0.70$\times$ & 9.12 & \textbf{51.4} \\
PaLU $0.5$ & 0.52$\times$ & 0.70$\times$ & 8.87 & 47.7 \\
\multicolumn{5}{@{}l}{\textit{Mistral-7B}} \\
\textbf{RAP} $0.7$ & \textbf{0.70}$\times$ & 0.70$\times$ & 5.69 & \textbf{50.9} \\
PaLU $0.5$ & 0.52$\times$ & 0.70$\times$ & 5.61 & 50.4 \\
\bottomrule
\end{tabular*}
\caption{Matched attention compute (${\approx}0.70\times$ of dense). At equal compute RAP keeps a larger share of the \KV{} cache and leads on LongBench on both models, while PaLU retains a perplexity edge of at most $0.25$. Bold marks the two axes the comparison is about.}
\label{tab:isoparam}
\end{table}

\noindent\textbf{End-to-end decode.}
At the operating point, full-model decode throughput relative to dense at batch~1 is $0.85\times$ at 4k, $0.96\times$ at 8k, $1.05\times$ at 16k and $1.11\times$ at 32k (Llama-3-8B, fp16, one A100-80GB PCIe, RAP on the fused Triton path against an SDPA dense reference).
The shape is the one the method predicts: what RAP shrinks is the cache, and the cache only dominates a decode step once the context is long, so below 8k the projection GEMMs still set the pace and RAP is behind.
We report one card and one kernel path only. The more compressed $\rho{=}0.5$ runs we have were measured on a different A100 variant, and at this granularity a hardware change is larger than the effect being measured, so we do not put the two in one table.
This is not the same effect as the attention-module measurement, and it took us a while to see that. At batch~1 the module ratio favors RAP up to 16k ($1.22$--$1.42\times$), then inverts at 32k ($0.91\times$), which is exactly where end-to-end throughput is at its best. So the end-to-end gain is a cache-residency effect, fewer bytes to read per step, and not a faster attention kernel.
One tension is visible from this appendix alone. The contexts where RAP wins on throughput are longer than the window this checkpoint was pretrained for, and \Cref{app:extended} reports that the dense model retrieves nothing at 16k. The two numbers measure different things, one whether a step is cache-bound and the other whether the output is useful. Read together they say that RAP's advantage lives in the cache-bound regime, and that reaching it usefully needs a model trained for that length.

\noindent\textbf{RoPE kernel microbenchmark.}
After RAP the retained pairs are no longer contiguous inside a head, so RoPE has to gather non-contiguous $\cos/\sin$ entries.
In PyTorch that gather materializes a new buffer before the rotation, and the copy can cost more than the rotation itself.
\Cref{fig:rope_kernel_heatmap} measures this on A100-80GB across sequence length and $\rho$, comparing Torch indexing and our fused Triton kernel against contiguous dense RoPE, with each cell split into a Torch triangle and a Triton triangle.
Torch often falls below $1\times$, which is the copy showing up as a slowdown, while Triton stays above $1\times$ and reaches about $4.6\times$ at long sequences.
This kernel is not where the headline speedups come from.
RoPE is a small fraction of a decode step, so the Prefill and Decode reductions in \Cref{fig:efficiency} come from dropping reconstruction and shrinking the projection GEMMs. The Triton path removes an implementation artifact of non-contiguous indexing. That is a statement about the kernel in isolation, and it cuts against us: end to end, the unfused path is in fact slightly faster at 16k and 32k. We do not claim the kernel as a source of the throughput numbers, only as the reason pair-aligned indexing costs nothing.
We report the microbenchmark because that artifact is specific to non-contiguous indexing and would otherwise be mistaken for a cost of pair-aligned pruning itself.

\noindent\textbf{Serving memory.}
\Cref{tab:serving} reports what the smaller cache buys at 8k context on one A100-80GB.
RAP caches $0.70\times$ the bytes per token dense does and peaks below dense at every batch we could run.
Both rows carry an implementation caveat, and they are not symmetric. PaLU caches low-rank latents and reconstructs $\K$ transiently, so its resident cache ought to be $\rho$ times dense. In the configuration that reaches its reported quality it instead measures the full $128$\,KiB per token, the same as dense, so the compression does not reach the cache at all there. Its peak is a separate matter: at 8k and batch~1 the resident cache is about a gigabyte either way, so the extra $15$\,GB PaLU carries over dense is transient allocation on the reconstruction path and not cache. We report peak memory because it is what decides the batch a server can run, and we do not decompose it further. The RAP column is likewise the better of the two implementations we have: our unfused variant peaks at $25.3$\,GB at batch~1, above dense, and runs out at batch~8, where dense still runs at batch~16.
So this table supports the narrow claim and not the broad one. RAP admits an implementation whose resident cache really is the retain ratio, and we show it. Whether PaLU admits none is not something these measurements establish.
The axis is capacity, not speed. This is a different, memory-optimized implementation, and it reaches $20.5$ tokens/s against $34.5$ for dense at batch~1. We make no throughput claim from it, and we do not attribute the gap to any one cause, because our own unfused end-to-end path reaches $30.5$ tokens/s on the same model.

\begin{table}[t]
\centering
\small
\setlength{\tabcolsep}{3pt}
\begin{tabular*}{\linewidth}{@{\extracolsep{\fill}}l c cccc@{}}
\toprule
& \textbf{B/tok} & \multicolumn{4}{c}{\textbf{Peak memory (GB)}} \\
\cmidrule(l){3-6}
\textbf{Method} & (KiB) & $B{=}1$ & $B{=}4$ & $B{=}8$ & $B{=}16$ \\
\midrule
Dense & 128 & 19.3 & 29.0 & 42.0 & 67.9 \\
PaLU $0.7$ & 128 & 34.8 & oom & oom & oom \\
\textbf{RAP} $0.7$ & \textbf{89.5} & \textbf{18.2} & \textbf{27.0} & \textbf{38.7} & \textbf{62.0} \\
\bottomrule
\end{tabular*}
\caption{Serving memory at 8k context on one A100-80GB (fp16). Bytes per token is the resident \KV{} state. ``oom'' marks batches at which PaLU ran out of memory.}
\label{tab:serving}
\end{table}

\end{document}